\documentclass[lettersize,journal]{IEEEtran}
\usepackage{amsmath,amsfonts}
\usepackage{algorithmic}
\usepackage{algorithm}
\usepackage{array}
\usepackage[caption=false,font=normalsize,labelfont=sf,textfont=sf]{subfig}
\usepackage{textcomp}
\usepackage{stfloats}
\usepackage{url}
\usepackage{verbatim}
\usepackage{multirow}
\usepackage{graphicx}
\usepackage{cite}
\usepackage{setspace}
\usepackage{colortbl}
\usepackage{makecell}
\usepackage{hyperref}
\hypersetup{hidelinks}

\begin{document}

\title{USLN: A statistically guided lightweight network for underwater image enhancement via dual-statistic white balance and multi-color space stretch}
\author{Ziyuan Xiao, Yina Han$^{\ast}$ \thanks{*Corresponding author}, Susanto Rahardja, and Yuanliang Ma
\thanks{The authors are with the School of Marine Science and Technology, Northwestern Polytechnical University, Xi'an 710072, China (e-mail: xiaoziyuan@mail.nwpu.edu.cn; yina.han@nwpu.edu.cn; susantorahardja@ieee.org; ylma@nwpu.edu.cn).}
}

\maketitle

\begin{abstract}
Underwater images are inevitably affected by color distortion and reduced contrast. Traditional statistic-based methods such as white balance and histogram stretching attempted to adjust the imbalance of color channels and narrow distribution of intensities a priori thus with limited performance. Recently, deep-learning-based methods have achieved encouraging results. However, the involved complicate architecture and high computational costs may hinder their deployment in practical constrained platforms. Inspired by above works, we propose a statistically guided lightweight underwater image enhancement network (USLN). Concretely, we first develop a dual-statistic white balance module which can learn to use both average and maximum of images to compensate the color distortion for each specific pixel. Then this is followed by a multi-color space stretch module to adjust the histogram distribution in RGB, HSI, and Lab color spaces adaptively. Extensive experiments show that, with the guidance of statistics, USLN significantly reduces the required network capacity (from existing $10^{6}$-$10^{8}$ to no more than $10^{3}$) and achieves state-of-the-art performance. The code and relevant resources are available at \url{https://github.com/deepxzy/USLN}.
\end{abstract}

\begin{IEEEkeywords}
underwater image enhancement, white balance, histogram stretching, neural network, deep learning.
\end{IEEEkeywords}

\section{INTRODUCTION}
\IEEEPARstart{U}{nderwater} images are of great significance for the utilization and research of underwater resources \cite{zhang2021enhancing,liu2020semantic}. However, underwater images have always been suffering from quality degradation. Specifically, since the red light travels the shortest in the water, the underwater images are always dominated by blue and green. Besides, the light is scattered by underwater micro suspending particles, which causes low contrast in underwater images. These problems seriously limit the ability to obtain the meaningful information that is not conducive to further processing of underwater images \cite{zhang2021enhancing}. Therefore, many algorithms \cite{UCM,RGHS,ULAP,Waternet,UWCNN,Ucolor,EUVP} are proposed to solve these issues.
\begin{figure}[!t]
\centering
\captionsetup[subfigure]{justification=centering}
\subfloat[Input]{\includegraphics[width=1.1in]{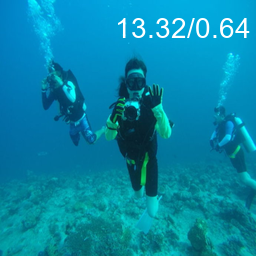}%
\label{fig_first_case}}
\hfil
\subfloat[Waternet\\*(1.09M)]{\includegraphics[width=1.1in]{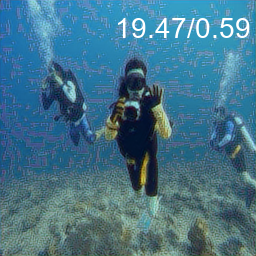}%
\label{fig_second_case}}
\hfil
\subfloat[FUnIE-GAN\\*(7.02M)]{\includegraphics[width=1.1in]{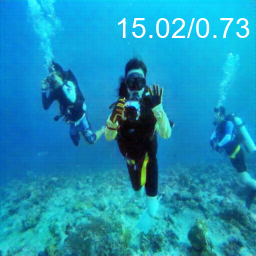}%
\label{fig_third_case}}
\hfil
\subfloat[UWCNN\\*(0.04M)]{\includegraphics[width=1.1in]{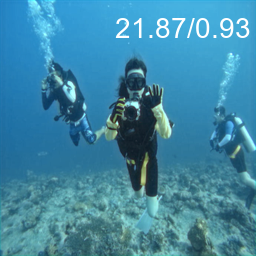}%
\label{fig_forth_case}}
\hfil
\subfloat[Ucolor\\*(148.77M)]{\includegraphics[width=1.1in]{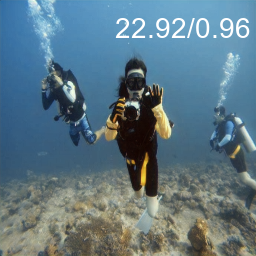}%
\label{fig_fifth_case}}
\hfil
\subfloat[Ours\\*(0.001M)]{\includegraphics[width=1.1in]{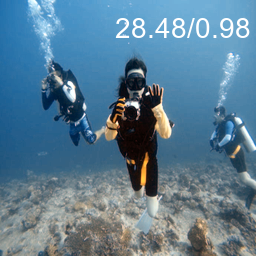}%
\label{fig_sixth_case}}
\caption{Intuitive comparisons on underwater images. The numbers below images represent the amount of parameters of different models. Our model removes color cast and improves contrast effectively. Compared with the existing methods, USLN takes the least parameters and achieves the best PSNR and SSIM scores.}
\label{fig_begin} 
\vspace{-1em}
\end{figure}

Researchers early focused on the degradation causes and pixel values of underwater images, proposing two targeted traditional methods, physical-model-based methods \cite{ARUIR,UDCP,tip2016,ULAP} and statistic-based methods \cite{ICM,UCM,ghani2015enhancement,ghani2017automatic}. Physical-model-based methods try to enhance images by reversing the underwater imaging physical model. Since the physical model is ill-posed, additional constraints must be introduced to estimate the relevant optical parameters. However, these artificial constraints do not always hold in changeable underwater environment, which affects the robustness of these methods. \cite{liu2020real}. In contrast, statistic-based methods mainly aim to improve the image quality directly by adjusting image pixel values a priori. For instance, the white balance-based methods \cite{li2020hybrid,ICM,finlayson2004shades} balance the image by multiplying the color channels by different handcrafted factors, while the histogram stretching-based methods \cite{CLAHE,ghani2015enhancement,ghani2017automatic} improve the contrast by stretching the color histogram of image from a narrower range to a wider range. These methods can improve images quality in color balance and contrast correction to a great extent. 
\IEEEpubidadjcol
However, the white balance-based methods multiply all pixel values in each color channel by the same factor, which ignores the different distortion degree of each pixel. For histogram stretching-based methods, because each color channel is stretched to a predefined range separately, these methods ignore the correlation between channels and introduce artificial enhancement effect inevitably. 

Recently, deep-learning-based methods \cite{pixelpixel,Waternet,Ucolor,UWCNN,EUVP} have been successfully used in this field. Early deep-learning-based methods reference existing noted backbones, which can directly learn the mapping relationship between raw image and clear image. However, as pointed by \cite{Ucolor}, these models ignore the domain knowledge of underwater image enhancement, which results in limited performance. To address this issue, Li et al. \cite{Waternet,Ucolor} took the results of traditional algorithms as additional inputs to enhance the response of networks towards specific regions. These methods exploit the advantages of traditional and deep-learning-based methods, which improve enhancement performance successfully. Nevertheless, existing deep-learning-based models \cite{pixelpixel,Waternet,Ucolor,UWCNN,EUVP} take numerous convolutional layers to extract high-dimensional image features. Thousands of parameters are involved to fit the underwater image enhancement mapping relationship, which makes these models complicate and time-consuming.

Based on above methods, we take a further step in this paper. We proposes a statistically guided lightweight network via dual-statistic white balance and multi-color space stretch. Specifically, dual-statistic white balance module embeds average and maximum of images to help model learn to compensate the color distortion for each specific pixel, while multi-color space stretch module adaptively increases the contrast of images by adjusting histogram distribution in different color spaces (RGB, HSI, Lab). As Fig. \ref{fig_begin} presented, compared with existing deep-learning-based methods, our model takes the least parameters (894) but achieves the best performance. The main contributions of this paper are highlighted as follows.
\begin{itemize}
\item{We propose two trainable modules, dual-statistic white balance module and multi-color space stretch module to remove color casts and improve contrast, respectively.}
\item{Our model follows the idea of statistic-based methods, that is modifying the pixel values in three-dimensional color space directly, rather than processing high-dimensional image features as current deep-learning-based methods do.}
\item{Our model takes the least parameters, but exceeds the SOTA models on several datasets (UIEB, Uimaginet, UFO).}
\end{itemize}

\section{RELATED WORK}
Underwater enhancement methods can be roughly classified into three categories: physical-model-based methods, statistic-based methods and deep-learning-based methods.

{\bf Physical-model-based methods.} Physical-model-based methods mainly focus on introducing visual priors to estimate the underwater optical parameters. Then, the clean images can be obtained by reversing the underwater imaging physical  model. These priors include red channel prior \cite{ARUIR}, underwater dark channel prior  \cite{UDCP}, minimum information prior \cite{tip2016}, underwater light attenuation prior \cite{ULAP}, and so on. Among them, underwater dark channel prior method \cite{UDCP} and red channel prior method \cite{ARUIR} both make improvement on the DCP algorithm \cite{DCP} for the underwater environment. UDCP only considers the blue and green channels, while ARUIR considers reverse red channel instead of red channel. 

Although these physical-model-based methods can achieve promising performance in some cases, these methods are sensitive to the types of underwater images, which means these priors are not always plausible in changeable and complicated underwater environment \cite{liu2020real}. Compared with these methods, our model can easily adapt to various underwater environments, due to the powerful learning capability of neural networks.

{\bf Statistic-based methods.} Statistic-based methods improve the images quality by adjusting pixel values. For example, Iqbal et al. proposed an unsupervised color correction method (UCM) \cite{UCM}. UCM firstly uses white balance to equalize the RGB channels to make image balanced. Then the histogram equalization is applied in the RGB and HSI color spaces respectively to increase the contrast of the images. On this basis, Ahmad et al. \cite{ghani2015enhancement} \cite{ghani2017automatic} proposed an adaptive histogram enhancement method base on limited Rayleigh distribution stretching, which can enhance details and reduce over-enhanced. These methods have obvious advantages for increasing the brightness and contrast of underwater images. 

These statistic-based methods need set many hyperparameters manually in advance. This always introduces artificial color. In contrast, our method integrates satistic-based methods into deep-learning-based methods, which automatically learns these hyperparameters according to different underwater environment.

{\bf Deep-learning-based methods.} Recently, deep-learning-based methods do impressive performance on underwater image enhancement task. These methods can be divided into convolutional neural networks (CNN) \cite{CNN} and generative adversarial networks (GAN) \cite{GAN}. For CNN-based models, Sun et al. \cite{pixelpixel} first proposed a pixel-to-pixel deep learning model based on encode-decoder framework, which has significant effect on underwater image denoising. Furthermore, Li et al. \cite{UWCNN} proposed a light-weight CNNs model (UWCNN) based on underwater scene prior, which can synthesize image datasets with different degrees of degradation. Most recently, inspired by physical-model-based methods, Li et al. \cite{Ucolor} proposed a model via medium transmission-guided multi-color space embedding (Ucolor), which uses reverse medium transmission map to emphasize quality-degraded regions. With regard to GAN-based model, Islam et al. \cite{EUVP} proposed FUnIE-GAN, which is inspired by U-net \cite{unet}, using five encoder-decoder pairs with skip-connection to build generator. The processing speed of this model is fast and results have a good effect on color recovery and sharpening. 

These deep-learning-based models need too much parameters to fit the mapping relationship between clear image and raw image. Even lightweight networks like UWCNN \cite{UWCNN} and Shallow-UWnet \cite{shallow} still contain thousands of parameters. In contrast, instead of extracting high-dimensional image features, USLN follows the idea of statistic-based algorithms, that is, to improve the quality of images by adjusting the pixel values in three color channels directly, thus greatly reducing the amount of parameters (merely takes 894 parameters).

\section{PROPOSED METHOD}
\begin{figure*}[!t]
\centering
\includegraphics[width=6in]{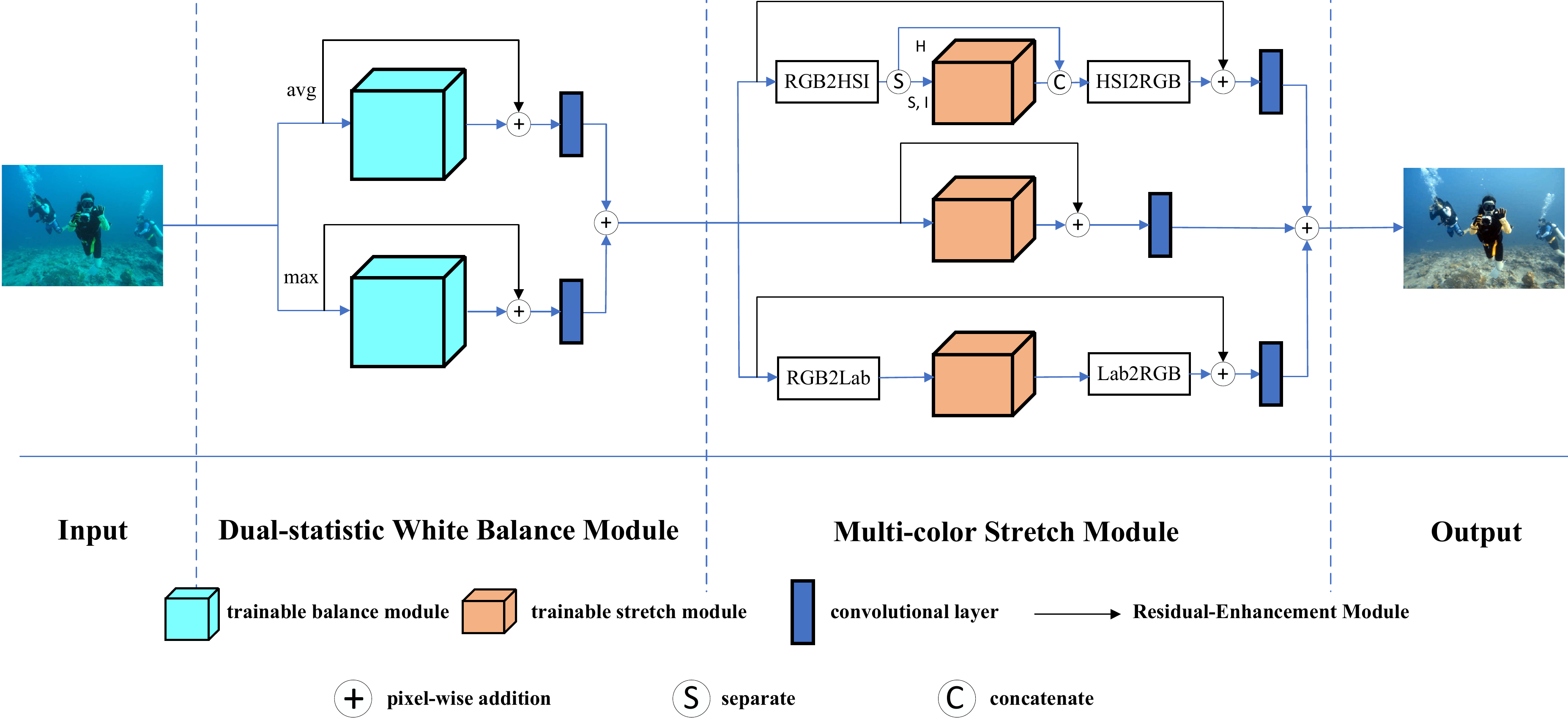}
\caption{Overview of the architecture of USLN. Our model consists of dual-statistic white balance module, multi-color space stretch module and several residual-enhancement modules. The input of USLN is three-dimensional underwater image in which the pixel values is between 0 and 1. ‘convolutional layer’ has the kernel of size 3 × 3 and stride 1, which is used to merge enhanced images together. A detailed network module structure is shown in follow subsections.}
\label{overview}
\vspace{-1em}
\end{figure*}

We present the overview architecture of USLN in Fig. \ref{overview}. USLN consists of two main modules, dual-statistic white balance module and multi-color space stretch module. {\bf In dual-statistic white balance module,} an underwater image is first fed into two trainable balance modules which are based on average and maximum of image, respectively. After getting the balanced images, there are also two residual-enhancement modules to recover details of images. Then, we merge the two enhanced images together and feed the final result to multi-color space stretch module. {\bf In multi-color space stretch module,} the image first goes through color space transformation where the RGB image is converted into HSI and Lab color spaces. We then get stretched images by feeding these multi-color spaces images into trainable stretch module, respectively. Simultaneously, in each color space, the original image is also forwarded to residual-enhancement module and pixel-wise add up to the stretched image. Finally, these enhanced images are all converted back to RGB color space and merge together. Note that in HSI color space, the H channel is split out and keep unchanged. 
 
In what follows, we detail all the components of our method, including dual-statistic white balance module (Sec. \uppercase\expandafter{\romannumeral3}-A), multi-color space stretch module (Sec. \uppercase\expandafter{\romannumeral3}-B), residual-enhancement module (Sec. \uppercase\expandafter{\romannumeral3}-C), and the loss function (Sec. \uppercase\expandafter{\romannumeral3}-D). In addition, we also discuss the superiority of our model from view of visual attention at the end of this section (Sec. \uppercase\expandafter{\romannumeral3}-E).

\begin{table*}
\caption{THE AMOUNTS OF PARAMETERS IN EACH MODULE OF USLN.}
\label{table}
\small
\setlength{\tabcolsep}{10pt}
\begin{spacing}{1.19}
\begin{tabular}{ccccc} 
\hline
Module                      & Dual-statistic White Balance Module        & Multi-color Space Stretching Module   &Residual-Enhancement Module  &Total\\
\hline
Parameters                   & 192       & 282     & 420   &894\\
\hline                                
\end{tabular}
\end{spacing} 
\label{tab1}
\vspace{-1em}
\end{table*}

\begin{figure}[!t]
\centering
\includegraphics[width=3.5in]{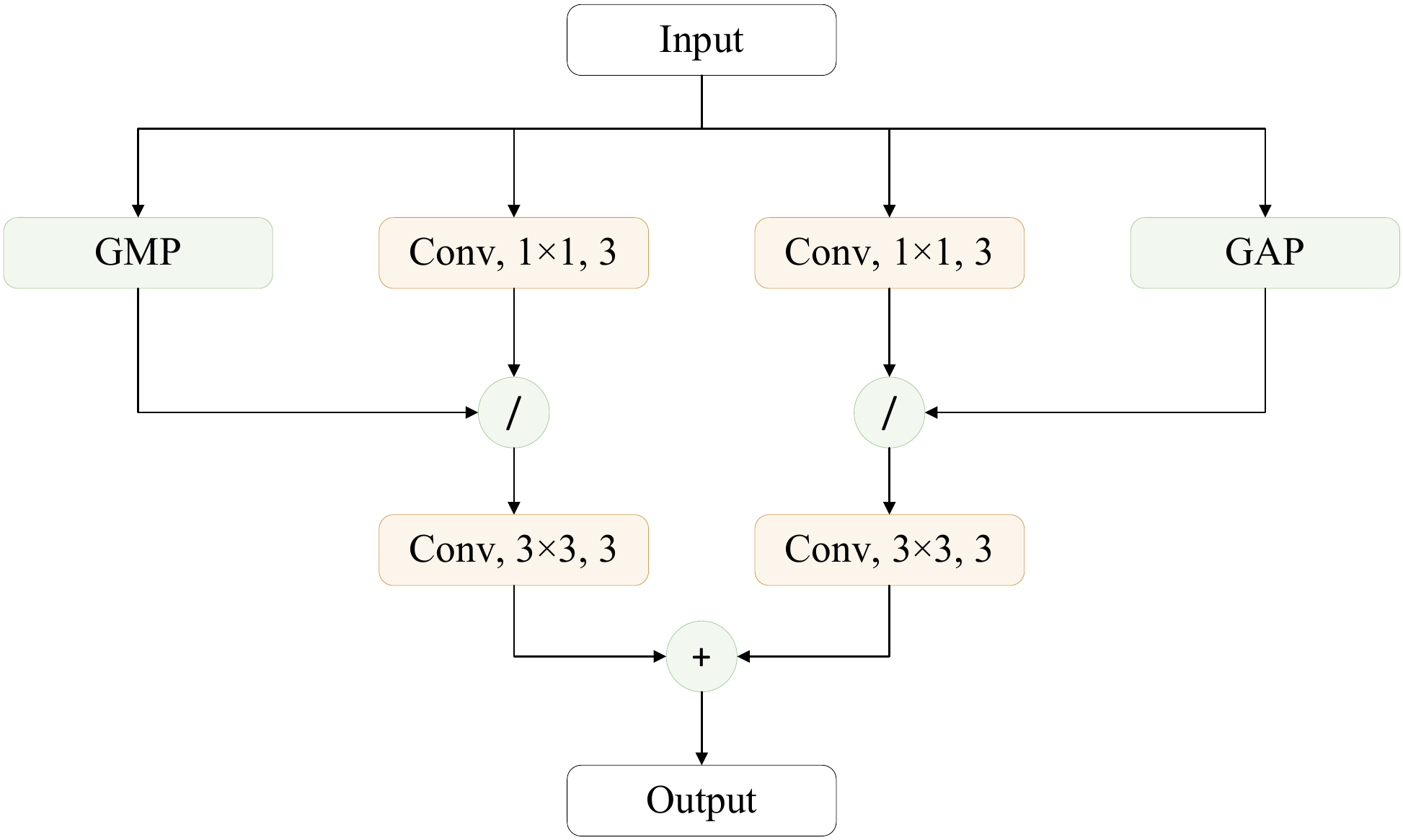}
\caption{The schematic illustration of dual-statistic white balance module, which consists of two trainable balance modules, the right part is based on Gray World theory, while the left part is based on White Patch algorithm. GMP and GAP are abbreviations of global max pooling and global average pooling, respectively.}
\label{wbm}
\vspace{-1em}
\end{figure}

\begin{figure}[!t]
\centering
\includegraphics[width=3.2in]{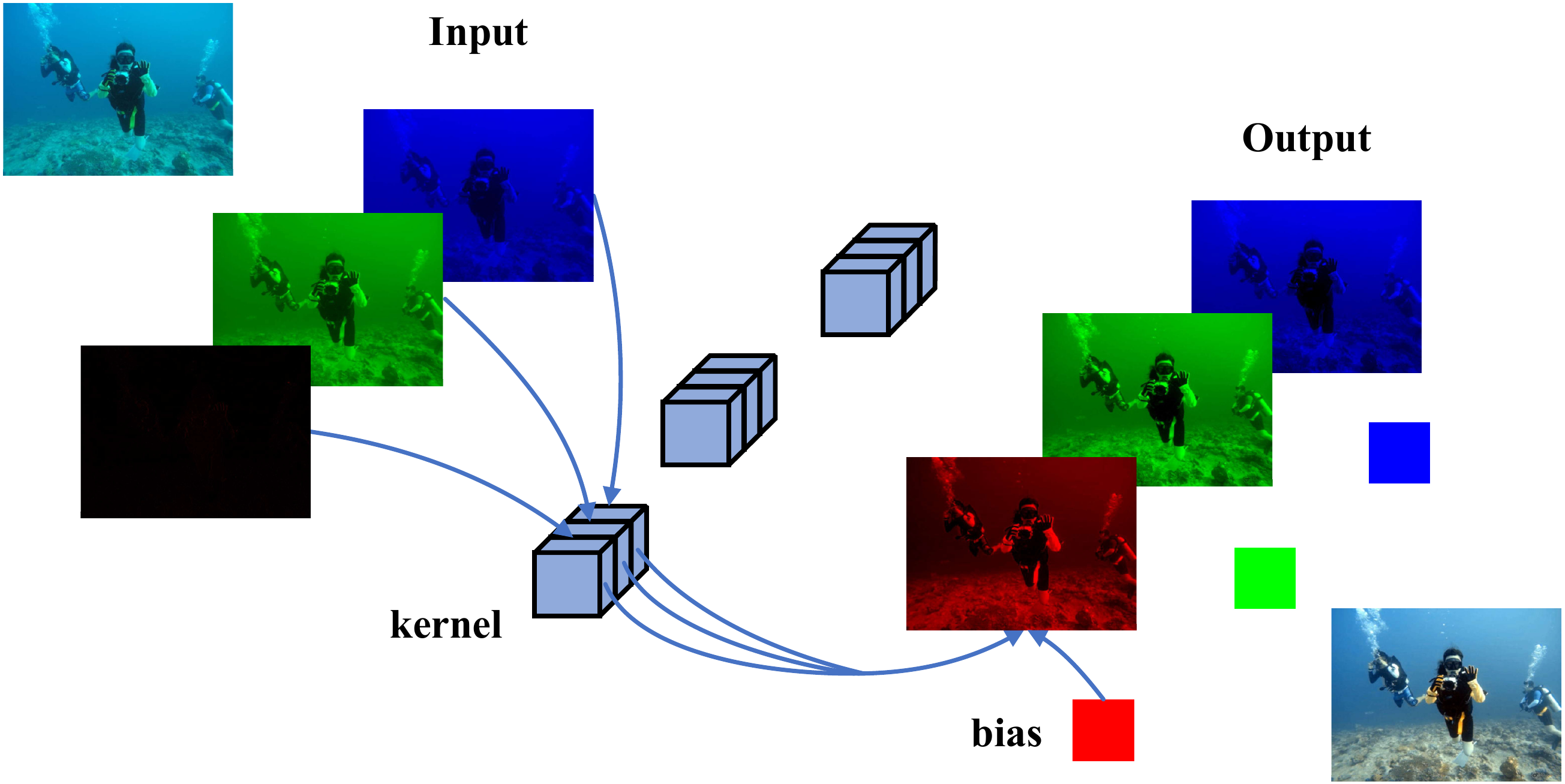}
\caption{The schematic illustration of pointwise convolution. We use pointwise convolution to generalize Von Kries hypothesis, which indicates not only the relationship between color channels but also correction of each channel.}
\label{1time1}
\vspace{-1em}
\end{figure}

\subsection{Dual-statistic Balance Module}

In the underwater environment, images are rarely color balanced. In general, there will be a phenomenon of higher blue and green light intensity and lower red light intensity. Hence, the first step in our model is to make image balanced. Inspired by white balance algorithm, we propose dual-statistic white balance module. 

The structure of dual-statistic white balance module is shown in Fig. \ref{wbm}. This module consists of two trainable balance modules, one is based on Gray World theory, while the other is based on White Patch algorithm. Next, we will explain the detailed derivation process of this module.

Von Kries hypothesis can be used to express the relationship between underwater images and ground-truth images:
\begin{equation}\label{1}
\setlength\abovedisplayskip{3pt}
\setlength\belowdisplayskip{3pt}
\left[\begin{array}{l}
I^{\prime}_{R} \\
I^{\prime}_{G} \\
I^{\prime}_{B}
\end{array} 
\right ]=
\left[ \begin{array}{ccc}
e_{R} & 0 & 0 \\
0 & e_{G} & 0 \\
0 & 0 & e_{B}
\end{array} 
\right ]
\left[\begin{array}{l}
I_{R} \\
I_{G} \\
I_{B}
\end{array}\right ]
\end{equation}
Where $I$ denotes pixels values of underwater images, while $I^{\prime}$ denotes those in ground-truth images, and $e_{R}$, $e_{G}$, $e_{B}$ denote scale factors that are applied independently to the three color channels.

However, Von Kries hypothesis considers the R, G, B color channels independently. More generally, we need to consider the relationship between color channels and correction of each channel, so as Fig. \ref{1time1} shown that we can use pointwise convolution to generalize Von Kries hypothesis as:
\begin{equation}\label{2}
\setlength\abovedisplayskip{3pt}
\setlength\belowdisplayskip{3pt}
I^{\prime} = Conv_{1 \times 1}(I)
\end{equation}
Where $Conv_{i}(\cdot)$ means the convolution operation, subscript i means the size of convolutional kernel.  

However, it is difficult to learn the relationship between underwater images and enhanced images only by pointwise convolution (Eq. (\ref{2})). Inspired by two white balance algorithms, Gray World and White Patch, we add average and maximum of images into Eq. (\ref{2}) to help model learn this relationship more easily.

The Gray World theory assumes that the average value of color object in a perfect image is gray, which means the average pixels of the three channels R, G, and B tend to the same value. Hence, scale factors of each channel, $e_{R}$, $e_{G}$, $e_{B}$ can be set based on GW theory:
\begin{align}\label{3}
\setlength\abovedisplayskip{3pt}
\setlength\belowdisplayskip{3pt}
A_{c}=\frac{1}{N} \sum_{i=1}^{N} I_{c}(x), c \in\{R, G, B\}
\end{align}
\begin{align}\label{4}
\setlength\abovedisplayskip{3pt}
\setlength\belowdisplayskip{3pt}
e_{R}=\frac{0.5}{A_{R}}, e_{G} &=\frac{0.5}{A_{G}}, e_{B}=\frac{0.5}{A_{B}} 
\end{align}
Where $A_{c}$ denotes the average value of $c$ channel in original image.

Therefore, inspired by Eq. (\ref{4}), we add average of each color channel in Eq. (\ref{2}), which can be expressed:

\begin{equation}\label{5}
\setlength\abovedisplayskip{3pt}
\setlength\belowdisplayskip{3pt}
I^{GW} = Conv_{1 \times 1}(I) \otimes \overline{A}
\end{equation}
where $\overline{A}=[\frac{1}{A_{R}},\frac{1}{A_{G}},\frac{1}{A_{B}}] \in \mathbb{R}^{3 \times 1}$, $\otimes$ denotes pixel-wise multiplication, $I^{GW}$ means image enhanced by the trainable white balance module based on Gray World algorithm. Gray World algorithm can be considered a special case of Eq. (\ref{5}).

White Patch algorithm assumes the maximum response of the RGB color channel is caused by the white patch in the scene. Theoretically, the white patch can reflect the color of the scene light, so the largest value in the RGB channel will be used as the light source. According to this hypothesis, scale factors of each channel can be expressed as:
\begin{align}\label{6}
\setlength\abovedisplayskip{3pt}
\setlength\belowdisplayskip{3pt}
{M}_{c}=\max I_{c}(x), c \in\{R, G, B\}
\end{align}
\begin{align}\label{7}
\setlength\abovedisplayskip{3pt}
\setlength\belowdisplayskip{3pt}
e_{R}=\frac{1}{M_{R}}, e_{G} &=\frac{1}{M_{G}}, e_{B}=\frac{1}{M_{B}} 
\end{align}
Where $M_{c}$ denotes the maximum value of $c$ channel in original image.

Same as before, inspired by Eq. (\ref{7}), we add maximum of each color channel in Eq. (\ref{2}), which can be expressed:

\begin{equation}\label{8}
\setlength\abovedisplayskip{3pt}
\setlength\belowdisplayskip{3pt}
I^{WP} = Conv_{1 \times 1}(I) \otimes \overline{M}
\end{equation}
where $\overline{M}=[\frac{1}{M_{R}},\frac{1}{M_{G}},\frac{1}{M_{B}}] \in \mathbb{R}^{3 \times 1}$, $I^{WP}$ means image enhanced by the trainable white balance module based on White Patch algorithm. White Patch algorithm can be considered a special case of Eq. (\ref{8}).

Finally, two enhanced image are merged together by going though 3×3 convolutional layer and pixel-wise adding together.
\begin{equation}\label{9}
\setlength\abovedisplayskip{3pt}
\setlength\belowdisplayskip{3pt}
I_{final} = Conv_{3 \times 3}(I^{GW}) \oplus Conv_{3 \times 3}(I^{WP})
\end{equation}
where $I_{final}$ means the final result enhanced by dual-statistic white balance module. $\oplus$ means pixel-wise addition.

\subsection{Multi-color Space Stretching Module}
Due to the low histogram range, underwater images often suffer from low contrast and visibility. Therefore, histogram stretching is adopted to provide a wider pixel distribution of the image channels to improve the image contrast. The following equation is used for linear contrast stretching:
\begin{equation}\label{10}
\setlength\abovedisplayskip{3pt}
\setlength\belowdisplayskip{3pt}
P_{out}=\frac{P_{in}-i_{min}}{i_{max}-i_{min}}(o_{max}-o_{min})+o_{min}
\end{equation}
Where the $P_{in}$ and $P_{out}$ are the input and output pixels intensity values, respectively, $i_{max}$ and $i_{min}$ represent the maximum and minimum intensity value of the input image, respectively, $o_{max}$ and $o_{min}$ represent the maximum and minimum intensity value of the desired output image, respectively.

In histogram stretching-based methods, $i_{max}$ and $i_{min}$ are easy to get, but $o_{max}$ and $o_{min}$ must be defined artificially. For example, Kashif Iqbal \cite{UCM} used global histogram stretching to enhance the contrast of the image, setting $o_{max}$, $o_{min}$ to 1, 0. However, such a simple setup may not achieve good results. We can use pointwise convolunsion to make Eq. (\ref{10}) trainable by next steps:

\begin{align}\label{11}
\setlength\abovedisplayskip{3pt}
\setlength\belowdisplayskip{3pt}
\widetilde{I}_{c}(x)=\frac{I_{c}(x)-\min I_{c}(x)}{\max I_{c}(x)-\min I_{c}(x)}, c \in\{R, G, B\}
\end{align}
\begin{align}\label{12}
\setlength\abovedisplayskip{3pt}
\setlength\belowdisplayskip{3pt}
I^{r} = Conv_{1 \times 1}(\widetilde{I})
\end{align}
Where $\widetilde{I}=[\widetilde{I}_{R},\widetilde{I}_{G},\widetilde{I}_{B}] \in \mathbb{R}^{3 \times H \times W}$, $I^{r}$ denotes the histogram stretched pixel value in RGB color space. Global histogram stretching can be considered a special case of Eq. (\ref{12}).

\begin{figure}[!t]
\centering
\includegraphics[width=3in]{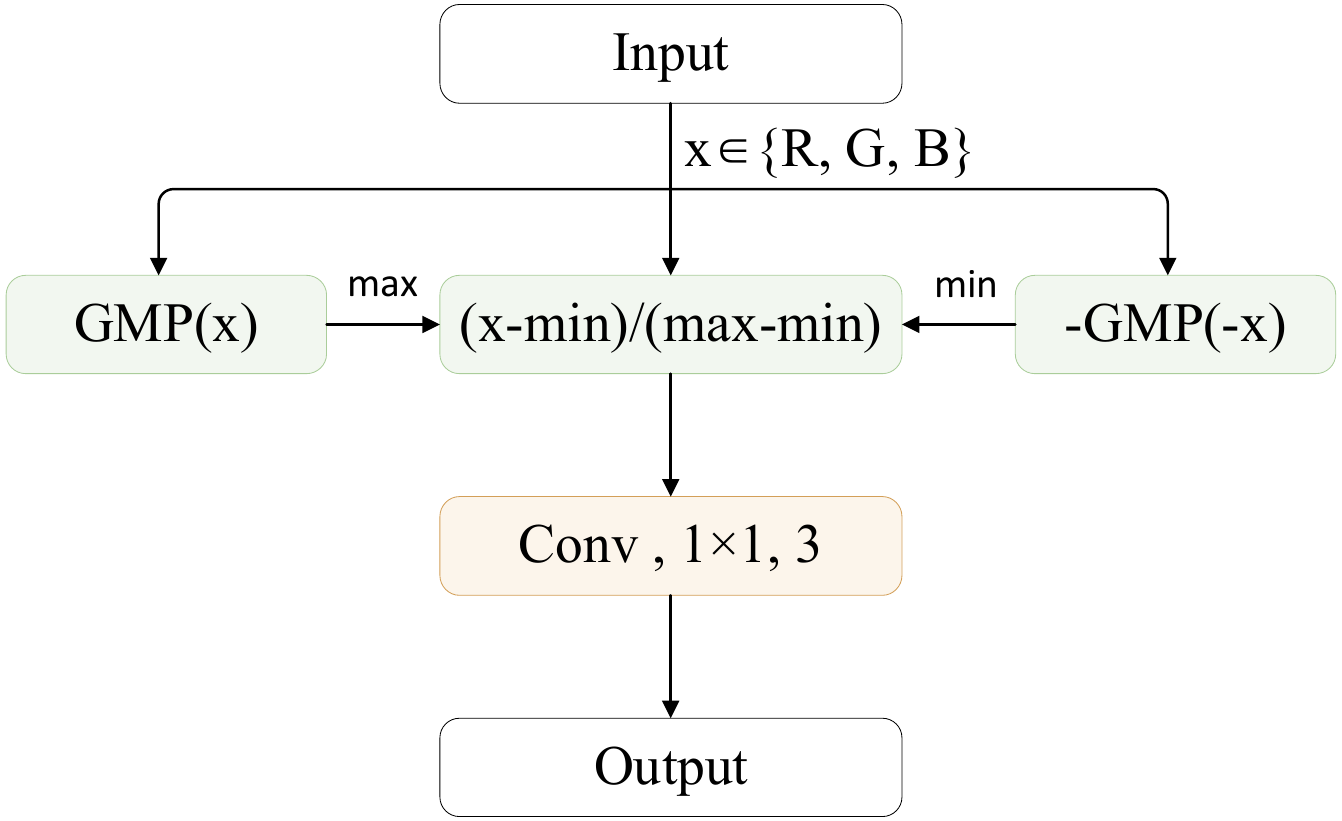}
\caption{The schematic illustration of trainable stretch module in RGB color space. GMP is the global max pooling. The left GMP is to get the maximum, while the right GMP is to get minimum.}
\label{HSRGB}
\vspace{-1em}
\end{figure}

Fig. \ref{HSRGB} presents the details of the trainable stretch module in RGB space.

\begin{figure*}[!t]
\centering
\subfloat[underwater image]{\includegraphics[width=1.5in]{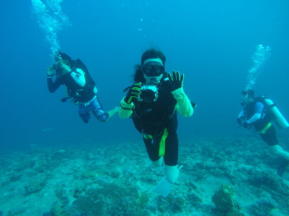}%
\label{raw}}
\hfil
\subfloat[RGB color space]{\includegraphics[width=1.65in]{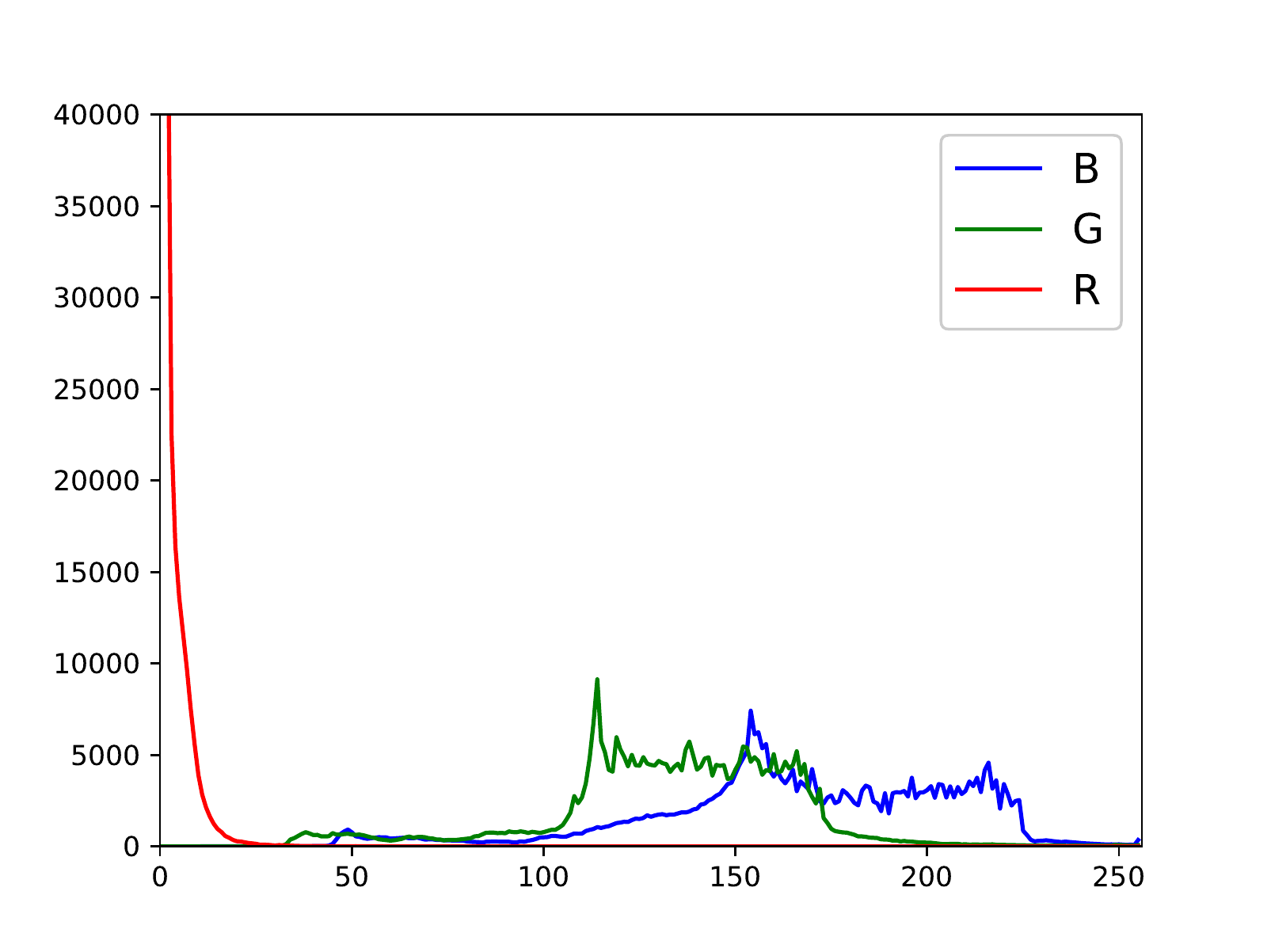}%
\label{fig_second_case}}
\hfil
\subfloat[HSI color space]{\includegraphics[width=1.65in]{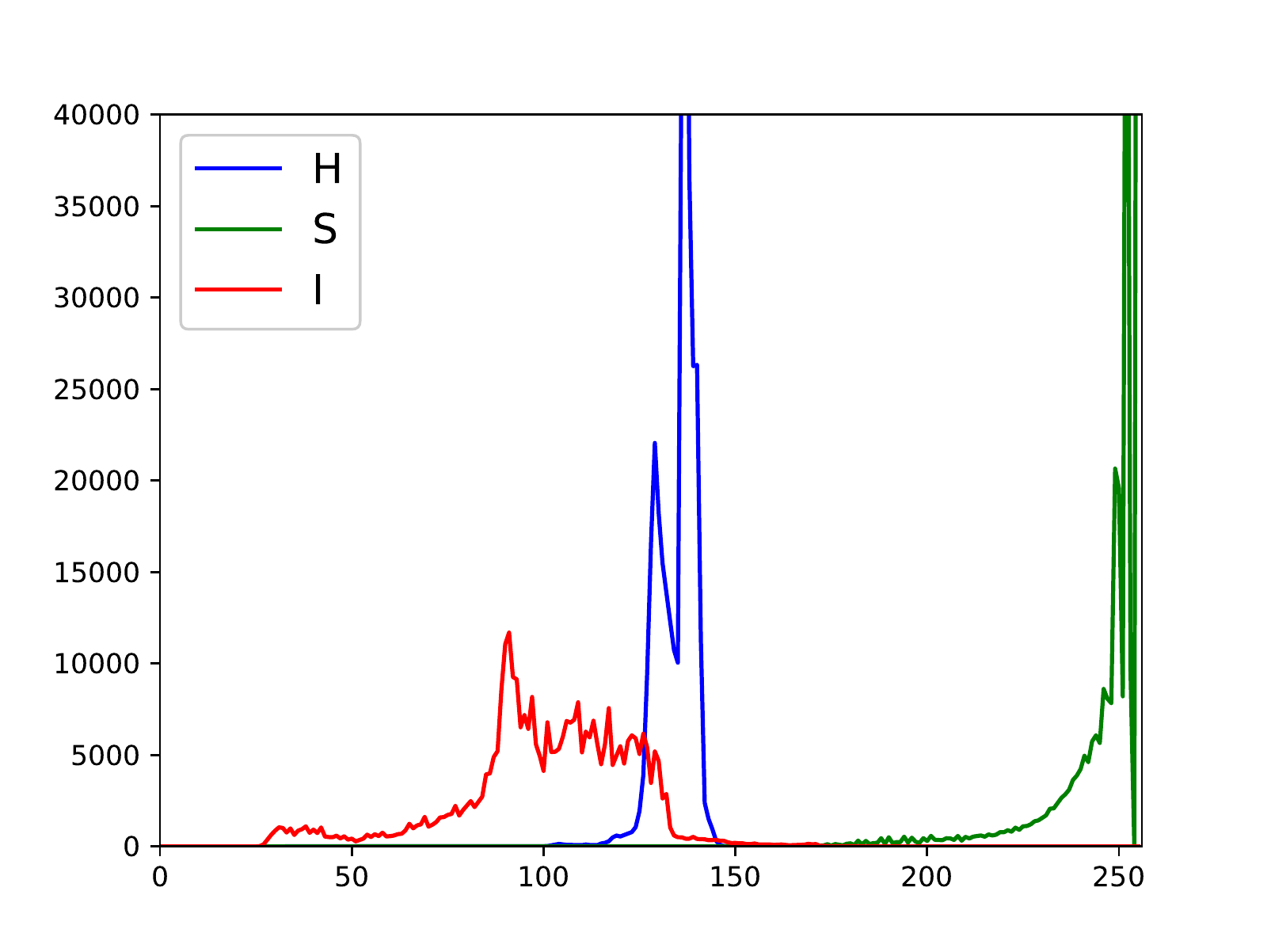}%
\label{fig_third_case}}
\hfil
\subfloat[Lab color space]{\includegraphics[width=1.65in]{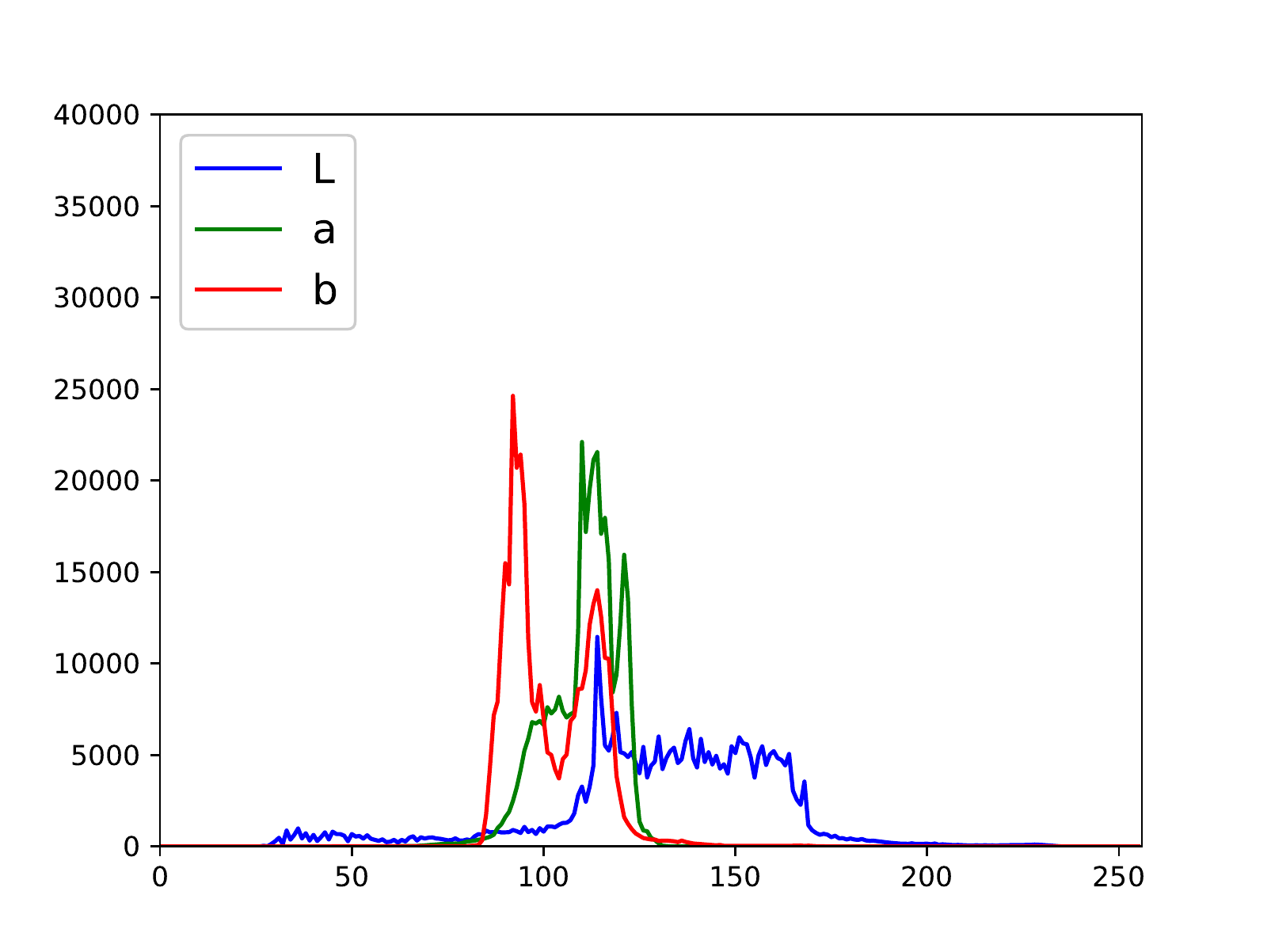}%
\label{fig_forth_case}}
\caption{An example of color distribution of underwater image (a) in different color spaces. (b), (c), (d) represent color distribution of (a) in RGB, HSI, Lab color spaces, respectively. }
\label{histogram}
\vspace{-1em}
\end{figure*}

In addition to operating in the RGB color space, inspired by traditional algorithms \cite{ICM,UCM,naik2003hue} and Ucolor \cite{Ucolor}, we also stretch images in HSI and Lab spaces. HSI color space reflects the way the human visual system perceives color, which intuitively reflects the hue, saturation and intensity. And Lab color space is designed to approach human perceptual vision, which contains almost every shade of every color in nature. As Fig. \ref{histogram} shown that the same image has different color distribution in various color spaces. Thus, the multiple color spaces stretch can facilitate the adjustment of underwater images color. In addition, the transformation relationship of images in these color spaces is not linear \cite{vgg}, which enables the deep-learning model to learn more complex mapping relationships. We also design ablation study to verity the contribution of multi-color space stretch operation.

Concretely, we convert the image into HSI and Lab color spaces and then stretch the S, I channels and L, a, b channels, respectively. Note that we do not stretch H channel, because our ablation studies verity that changing the value of the H channel will damage the color of the image. Interestingly, this conclusion is consistent with the statistic-based methods like \cite{UCM}, \cite{ICM} and \cite{abdul2014underwater}.

At last, three stretched images are converted back to RGB color space and merged together by going though 3×3 convolutional layer and pixel-wise add up.
\begin{equation}\label{13}
\setlength\abovedisplayskip{3pt}
\setlength\belowdisplayskip{3pt}
I_{final} = Conv_{3 \times 3}(I^{r}) \oplus Conv_{3 \times 3}(I^{h}) \oplus Conv_{3 \times 3}(I^{l})
\end{equation}
Where $I^{r}$, $I^{h}$, $I^{l}$ denote the histogram stretched pixel value in RGB, HSI, Lab color spaces, respectively.

It should be noted that we made a pytorch implementation that converts image between RGB, HSV, and Lab color spaces allowing differentiable back-propagate, so our model is trained in an end-to-end manner.

\subsection{Residual-Enhancement Module}
\begin{figure}[!t]
\centering
\includegraphics[width=3in]{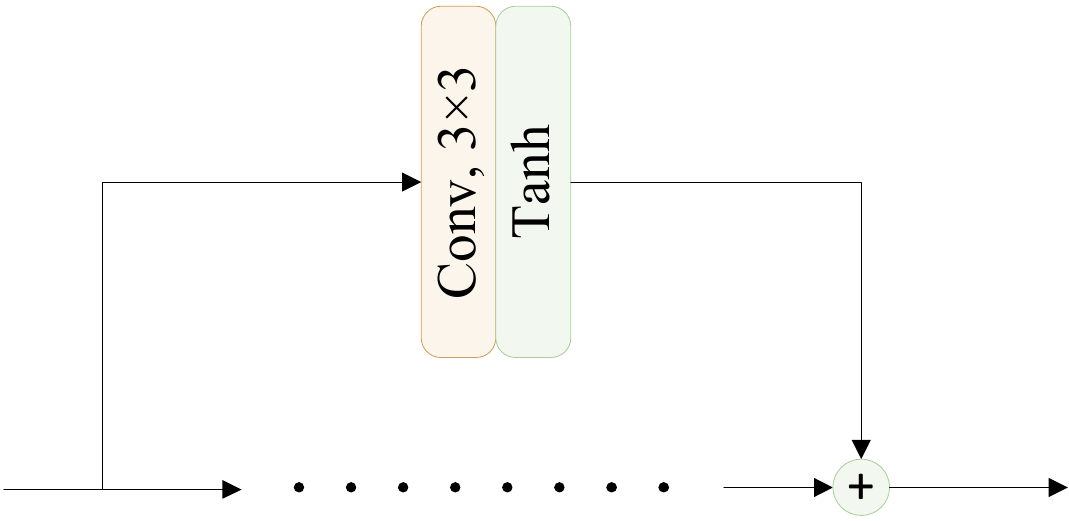}
\caption{The schematic illustration of residual-enhancement module consists of 3×3 convolutional layer and Tanh activation function.}
\label{RES}
\vspace{-1em}
\end{figure}

Drawing on deep learning techniques, we introduced residual enhancement module to our model, which aims to preserve the data fidelity and increase capacity of model \cite{res}. Fig. \ref{RES}. shows the structure of residual-enhancement module, which consists of a convolutional layer and an activation function. 

\begin{equation}
\setlength\abovedisplayskip{3pt}
\setlength\belowdisplayskip{3pt}
res = \delta\left(Conv_{3 \times 3}(z)\right)
\end{equation}
where $z$ and $res$ mean the input and output of residual-enhancement module, respectively, $\delta(\cdot)$ represents the Tanh activation function.

The kernel size of convolutional layer is set as 3×3, because the main convolutional layers in dual-statistic white balance module and multi-color space stretch module are pointwise convolution, which ignores the relationship between adjacent pixels. Therefore, the 3×3 convolutional layers in residual-enhancement module can alleviate this problem. For activation function, we chose Tanh activation function, because the output of Tanh activation is between -1 and 1, which means that residual-enhancement module can choose to reduce or increase the intensity of the enhanced pixels based on the features.

\subsection{Loss Function}
Inspired by Hang Zhao’s work \cite{zhao2016loss} and Johnson’s work \cite{johnson2016perceptual}, we used combined loss to train our model, which consists of MAE loss, SSIM loss and VGG perceptual loss.

MAE loss is to calculate the mean of the absolute difference between all pixels of the output image, $\hat{y}$, and the label, $y$, it can be expressed as:
\begin{equation}
\setlength\abovedisplayskip{3pt}
\setlength\belowdisplayskip{3pt}
L_{MAE}=\sum_{m=1}^{W}\sum_{n=1}^{H}\left|\hat{y}(m, n)-y(m, n)\right|
\end{equation}

SSIM Loss is to improve the structural similarity between the feature representations of the output image, $\hat{y}$, and the label, $y$:
\begin{equation}
\setlength\abovedisplayskip{3pt}
\setlength\belowdisplayskip{3pt}
L_{SSIM}=1-\frac{\left(2 u_{\hat{y}} u_{y}+\mu_{1}\right)\left(2 \sigma_{\hat{y} y}+\mu_{2}\right)}{\left(u_{\hat{y}}^{2}+u_{y}^{2}+\mu_{1}\right)\left(\sigma_{\hat{y}}^{2}+\sigma_{y}^{2}+\mu_{2}\right)}
\end{equation}
where u and $\sigma$ denote the mean, standard deviation, and covariance of images. $\mu_{1}$ and $\mu_{2}$ are used to stabilize the division.

VGG perceptual loss is based on pretrained VGG networks. Let $\phi(\cdot)$ be the last convolutional layer of VGG-19 networks \cite{vgg}. The perceptual loss is calculated the MAE loss of the feature representations of output image, $\hat{y}$, and the label, $y$, assuming that the total number of pixels is $M$:
\begin{equation}
\setlength\abovedisplayskip{3pt}
\setlength\belowdisplayskip{3pt}
L_{VGG}=\sum_{m=1}^{W}\sum_{n=1}^{H}\left|\phi(\hat{y})(m, n)-\phi(y)(m, n)\right|
\end{equation}

The final combined loss is the linear combination of the MAE loss, SSIM loss and VGG perceptual loss:
\begin{equation}
\setlength\abovedisplayskip{3pt}
\setlength\belowdisplayskip{3pt}
L_{final}=L_{M A E}+\lambda_{1} L_{S S I M}+\lambda_{2} L_{V G G}
\end{equation}
where $\lambda_{1}$ and $\lambda_{2}$ are empirically set to 0.25 and 1, respectively, for balancing the scales of different losses.

The validity of our combined loss function will be demonstrated in later ablation experiments.

\subsection{Perspective of Attention Mechanism}
From perspective of attention mechanism, we think that our model can be seen as a special case of channelwise attention networks. Classic channelwise attention networks, SE block  \cite{SE} can be roughly divided into three stages: squeeze, excitation and scale. SE block first squeezes global spatial information into a channel descriptor, then the global information is embedded by channelwise multiplying the excited channel descriptor and feature map together. In our model, the squeezed channel descriptors include average, maximum, and minimum of images, then they are excited by Eq. (\ref{5}), Eq. (\ref{8}), and Eq. (\ref{12}), respectively. These operations can exploit the global information effectively.

\section{EXPERIMENTS}
In this section, we first describe the datasets and evaluation metrics, then introduce our implementation details. We compare our model with the current underwater image enhancement models and provide ablation experiments to verify the contribution of each module. Cases of failure are presented at the end of this section.

\subsection{Benchmarks}
In our work, we adopted four benchmarks for evaluation, Uimagine \cite{EUVP}, UIEB \cite{Waternet}, UFO \cite{UFO}, and Color-Check7 \cite{colorcheck}. The scale of these datasets is various, so that we can compare these methods comprehensively.

In UIEB \cite{Waternet}, Uimagine \cite{EUVP}, and UFO \cite{UFO}, there are many paired images and unpaired images which were divided as shown in Tab. \uppercase\expandafter{\romannumeral2}.

Color-Check7 \cite{colorcheck} contains seven original underwater images with Color Checker. These images are taken by seven different cameras, Canon D10, Fuji Z33, Olympus Tough 6000, Olympus Tough 8000, Pentax W60, Pentax W80, and Panasonic TS1, denoted as D10, Z33, T6000, T8000, W60, W80, TS1. Considering there is no corresponding train set in Color-Check7, we got three results from models trained on Train-800, Train-3145 and Train-1500, respectively. Finally, the best score is used to evaluate the performance. 

\subsection{Evaluation Metrics}
For test sets in UIEB, Uimagine and UFO, we chose three full-reference evaluation metrics and one no-reference evaluation metric. Concretely, 1) The full-reference evaluation metrics consist of Peak Signal-to-Noise Ratio (PSNR) \cite{psnrssim}, Structural Similarity (SSIM) \cite{psnrssim} and Mean Squared Error (MSE), which are used for paired test sets (Test-90, Test-555, Test-120). A higher PSNR or a lower MSE score means that the output image and the label image are closer in content, while a higher SSIM score means that the two images are more structurally similar. 2) The no-reference evaluation metric is UIQM \cite{UIQM}, which is used for unpaired test sets (Test-60, Test-1270). A higher UIQM score suggests a better human visual perception. UIQM is the linear combination of UICM, UISM, and UIConM, which reflect the degree of color deviation, brightness, and contrast, respectively. However, it is worth mentioning that these no-reference metrics cannot accurately reflect the quality of an image in some cases \cite{Waternet} \cite{berman2020underwater}, so scores of UIQM (UICM, UISM, UIConM) are only provided as references for our study. We will present enhanced unpaired images in Visual Comparisons section for readers to score.

For Color-Check7, each colorcheck in Color-Check7 consists of 24 color patches. We used CIEDE2000 \cite{CIEDE} to evaluate the enhanced results. Concretely, we cropped the 24 color patches for each enhanced result and measured the average value of each color patch, then calculated the perceptual differences between corresponding color patches of ground-truth Color Checker and those of enhanced results.

\begin{table}
\centering
\centering
\caption{THE PRESENTATION OF DATASET DIVISION. THE NUMBLES REPRESENTS THE AMOUNT OF IMAGES IN SETS}
\label{table}
\small
\setlength{\tabcolsep}{10pt}
\begin{spacing}{1.19}
\begin{tabular}{l|l|l|l}
\hline
\multirow{2}{*}{ } & \multirow{2}{*}{Train set} & \multicolumn{2}{c}{Test set} \\
\cline{3-4}
                  &                            & paired       & unpaired      \\
\hline
UIEB \cite{Waternet}             & Train-800                  & Test-90      & Test-60       \\
\hline
Uimagine \cite{EUVP}         & Train-3145                 & Test-555     & Test-1270    \\
\hline
UFO \cite{UFO}        & Train-1500                 & Test-120     & None    \\
\hline
\end{tabular}
\end{spacing}
\label{tab2}
\vspace{-1em}
\end{table}

\subsection{Implementation Details}
We used python and pytorch framework via NVIDIA RTX3080 on windows 10. Adam was used for network optimization, while the initial learning rate was set as 0.01 with 5\% reduction per epoch. We trained our model for 100 epochs and the batch size was set as 10. 

Different from fixed size (256×256) of images in Uimagine, the size of images in UIEB is unfixed. In order to use batch learning, all images in UIEB were resized to 256×256. For UFO, the raw images are low-resolution (320×240), while the ground truth images are high-resolution (640×480). However, our research in this paper has nothing to do with the super-resolution task, so all images in UFO were also resized to 256×256. 

Note that except for resizing images, we did not pre-process the images in any other way in our experiments.

\subsection{Compared Methods}

We selected eight works to make fair comparisons with our model, including two statistic-based methods (UCM \cite{UCM}, RGHS \cite{RGHS}), two physical physical-model-based methods (Li et al. \cite{tip2016}, ULAP \cite{ULAP}), four deep-learning-based methods (Waternet \cite{Waternet}, Ucolor \cite{Ucolor}, UWCNN \cite{UWCNN}, FUnIE-GAN \cite{EUVP}). UCM integrates White Balance and histogram stretch to improve image quality, which is highly similar to our model. The comparison with UCM aims to demonstrate the advantages of our trainable modules.

For these traditional unsupervised methods, we used them directly on the test sets. For these deep-learning-based methods, we used the code and training strategy provided by the authors. Note that although FUnIE-GAN can be trained with unpaired data, we also trained this model in paired train sets to ensure the fairness of the experiment.

In addition to the quality of the enhancement results, for deep-learning-based models, GFLOPs and the amount of parameters are also important evaluation metrics. As shown in Table \uppercase\expandafter{\romannumeral3}, our model takes the least parameters and GFLOPs. Compared with the second-lightest model, the amount of parameters and GLOPs of our model are only 0.22\% of UWCNN (amount of parameters) and 7.89\% of FUnIE-GAN (GFLOPs). 

\begin{table}
\caption{THE MODEL GFLOPS AND PARAMETERS (IMAGE SIZE IS 256×256, BATCHSIZE IS 1)}
\label{table}
\small
\setlength{\tabcolsep}{20pt}
\begin{spacing}{1.19}
\begin{tabular}{ccc} 
\hline
Models                      & Parameters        & GFLOPs       \\
\hline
USLN                   & $8.49 \times 10^{2}$       & 0.06       \\
\hline
Waternet \cite{Waternet}                   & $1.09 \times 10^{6}$       & 142.91     \\
\hline
UWCNN \cite{UWCNN}             & $4.00 \times 10^{4}$       & 0.79            \\
\hline
FUnIE-GAN \cite{EUVP}             & $7.02 \times 10^{6}$       & 0.73         \\
\hline
Ucolor \cite{Ucolor}             & $1.49 \times 10^{8}$       & 2805.34         \\
\hline                                
\end{tabular}
\end{spacing} 
\label{tab3}
\vspace{-1em}
\end{table}

\subsection{Quantitative Comparisons}
We first make a quantitative comparison on Test-90, Test-555, and Test-120. The average scores of MSE, PSNR, and SSIM of different methods are shown in Table \uppercase\expandafter{\romannumeral4}. As shown in Table \uppercase\expandafter{\romannumeral4}, although there are only few parameters in our model, our method still performs better than the complex deep-learning-based models. Our model achieves the lowest MSE score and highest PSNR, SSIM scores in all three datasets. In addition, we also found that 1) although our method principle is roughly the same as UCM model, which is based on white balance and histogram stretch, our model performs much better than UCM, which suggests the superiority of our trainable modules. 2) Our model shows excellent performance on small datasets UIEB. Compared with the second-best performer, our USLN achieves the percentage gain of 16\%, 8\% and 5\% in terms of MSE, PSNR and SSIM, respectively. This means that our model can learn the mapping of underwater images and enhanced images well with small dataset.

Then, for Color-Checker7, we show the average CIEDE2000 scores of different methods in Table \uppercase\expandafter{\romannumeral5}. Our model achieves highest color similarity in Z33, T6000 and T8000. Moreover, our model achieve the lowest average CIEDE2000 score (10.71 under CIEDE2000 metric). These results show that our model is robust to different cameras and can restore the real color of underwater images with high accuracy. Interestingly, many compared methods get worse average CIEDE2000 score than raw images, which suggests that these methods even destroy the images color.

Next, we use UIQM as evaluation metric for unpaired test set (Test-60, Test-1270). The average scores of different methods are shown in Table \uppercase\expandafter{\romannumeral5}. As Table \uppercase\expandafter{\romannumeral6} shown, Our model gets highest UICON in Test-90, and second best UICM and UICON scores in Test-1270. There are two interesting findings from the Table \uppercase\expandafter{\romannumeral5}, 1) Different from full-reference evaluation metrics, deep-learning-based models have not achieve much better results than traditional methods in no-reference evaluation metrics. Conversely, for UICM, Li et. al and UCM rank the best and second best respectively in both datasets. 2) Although FUnIE-GAN and Ucolor perform best in Test-90 and Test-1270, respectively, they do not perform well in another dataset. In comparison, our model achieves robust performance in both two test sets.

\begin{table*}
\centering
\caption{UNDERWATER IMAGE ENHANCEMENT PERFORMSNCE METRIC IN TERMS OF AVERAGE MSE, PSNR AND SSIM VALUES. WE REPRESENT THE BEST TWO RESULTS IN \textcolor{red}{RED} AND \textcolor{blue}{BLUE} COLORS.}
\label{table}
\small
\begin{spacing}{1.19}
\setlength{\tabcolsep}{10pt}
\begin{tabular}{c|ccc|ccc|ccc} 
\hline
\multirow{2}{*}{model} & \multicolumn{3}{c|}{Test-90 (UIEB)} & \multicolumn{3}{c|}{Test-555 (Uimagine)} & \multicolumn{3}{c}{Test-120 (UFO)} \\
\cline{2-10}
                       &MSE $\downarrow$ & PSNR $\uparrow$      & SSIM $\uparrow$    &MSE $\downarrow$   & PSNR $\uparrow$       & SSIM $\uparrow$  &MSE $\downarrow$   & PSNR $\uparrow$       & SSIM $\uparrow$\\
\hline
raw       & 2037 & 16.84   & 0.76  & 1608 & 16.71 & 0.72 & 708 & 20.41& 0.76 \\
\hline
UCM \cite{UCM}                   & 1775       & 17.52      & 0.82      & 856 & 19.54       & 0.76      & 1385     & 18.41 & 0.74\\
\hline
RGHS \cite{RGHS}              & 1261       & 19.12      & 0.83     & 1694 & 16.36       & 0.70      & 1283     & 17.79 & 0.72\\
\hline
ULAP \cite{ULAP}               & 2395      & 15.95      & 0.75      & 1202 & 18.12       & 0.68     & 979      & 19.09  & 0.71\\
\hline
Li et al. \cite{tip2016}              & 1849       & 18.18       & 0.77       & 2157 & 15.97       & 0.64      & 2722     & 14.73  & 0.61\\
\hline
Waternet \cite{Waternet}               & 1219      & 19.03       & 0.79       & 559 & 21.53       & 0.79      & 524      & 21.66 & 0.77\\
\hline
Ucolor \cite{Ucolor}            & \textcolor{blue}{572}       & \textcolor{blue}{21.95}       & \textcolor{blue}{0.88}      & \textcolor{blue}{317}  & \textcolor{blue}{23.85}   & 0.80& \textcolor{blue}{286}  & \textcolor{blue}{24.37}  & 0.78\\
\hline
FUnIE-GAN \cite{EUVP}              & 1476       & 17.75      & 0.78     & 813 & 20.14       & 0.68      & 418  & 22.49   & 0.67\\
\hline
UWCNN \cite{UWCNN}           & 1126      & 19.22       & 0.87     & 351 & 23.54       & \textcolor{blue}{0.82}      & 350   & 23.55   & \textcolor{blue}{0.80} \\
\hline
USLN       & \textcolor{red}{479} & \textcolor{red}{23.78}   & \textcolor{red}{0.92}  & \textcolor{red}{311} & \textcolor{red}{24.13} & \textcolor{red}{0.83}& \textcolor{red}{280}& \textcolor{red}{24.54}& \textcolor{red}{0.81} \\
\hline
\end{tabular}
\end{spacing}
\label{tab4}
\vspace{-1em}
\end{table*}

\begin{table*}
\centering
\caption{UNDERWATER IMAGE ENHANCEMENT PERFORMSNCE METRIC IN TERMS OF AVERAGE UICM, UISM, UICONM AND UIQM VALUES. WE REPRESENT THE BEST TWO RESULTS IN \textcolor{red}{RED} AND \textcolor{blue}{BLUE} COLORS.}
\label{table}
\small
\begin{spacing}{1.19}
\setlength{\tabcolsep}{12pt}
\begin{tabular}{c|cccc|cccc} 
\hline
\multirow{2}{*}{model} & \multicolumn{4}{c|}{Test-60 (UIEB)} & \multicolumn{4}{c}{Test-1270 (Uimagine)}  \\
\cline{2-9}
                       & UIQM $\uparrow$         & UICM $\uparrow$       & UISM $\uparrow$         & UICON $\uparrow$   & UIQM $\uparrow$         & UICM $\uparrow$      & UISM $\uparrow$        & UICON $\uparrow$ \\
\hline
raw                   & 2.10       & 3.10       & 3.97      & 0.24   & 2.43       & 5.49       & 5.95      & 0.15\\
\hline
UCM \cite{UCM}                   & 2.21       &\textcolor{blue}{6.40}      & 4.20      & 0.22   & 2.72       &\textcolor{blue}{8.71}      & 6.44      & 0.16\\
\hline
RGHS \cite{RGHS}              & 2.06       & 5.78       & 3.98       & 0.20  & 2.12       &7.07      & 5.19      & 0.11 \\
\hline
ULAP \cite{ULAP}                & 1.55      & 5.17       & 3.07       & 0.14  & 2.21       &7.24      & 5.64      & 0.10 \\
\hline
Li et al. \cite{tip2016}               & 2.26       & \textcolor{red}{8.80}       & 4.10       & 0.22   & 2.80       &\textcolor{red}{9.36}      & 6.54      & 0.17\\
\hline
Waternet \cite{Waternet}                &\textcolor{blue}{2.76}      & 4.33       & \textcolor{blue}{5.54}       & 0.28   & 2.80       &4.98      &\textcolor{blue}{6.55}      & 0.21\\
\hline
Ucolor \cite{Ucolor}             & 2.72       & 4.42       & 5.48      & 0.28  & \textcolor{red}{3.11}       &4.35      & \textcolor{red}{6.78}      &\textcolor{red}{0.28} \\
\hline
FUnIE-GAN \cite{EUVP}              & \textcolor{red}{2.83}       & 5.66       & \textcolor{red}{6.59}      & 0.20   & 2.58       &6.22      & 6.40      & 0.15\\
\hline
UWCNN \cite{UWCNN}            & 2.63       & 3.23       & 5.06      & \textcolor{blue}{0.29}  & 2.81       &4.53     & 6.41      & 0.22 \\
\hline
USLN                   & 2.52       & 4.38       & 4.50      & \textcolor{red}{0.30}   & \textcolor{blue}{2.86}       & 4.53       & 6.47      & \textcolor{blue}{0.23}\\
\hline
\end{tabular}
\end{spacing}
\label{tab5}
\vspace{-1em}
\end{table*}

\begin{table*}
\centering
\caption{THE COLOR DISSIMILARITY COMPARISONS OF DIFFERENT METHODS ON COLOR-CHECK7 IN TERMS OF THE CIEDE2000. WE REPRESENT THE BEST TWO RESULTS IN \textcolor{red}{RED} AND \textcolor{blue}{BLUE} COLORS.}
\label{table}
\small
\setlength{\tabcolsep}{15pt}
\begin{spacing}{1.19}
\begin{tabular}{ccccccccc} 
\hline
Models                      & D10        & Z33     & T6000        & T8000  & TS1        & W60 & W80       & {\bf Avg}   \\
\hline
raw                   & 11.62      & 16.00    & 13.37      & 16.44  & 14.46      & 10.84 & 9.80      & {\bf 13.27}   \\
\hline
UCM \cite{UCM}                  & 12.23    & \textcolor{blue}{10.87}    & 12.32     & \textcolor{red}{11.79}  & 12.75      & 11.99  & \textcolor{red}{8.04}      & {\bf \textcolor{blue}{11.43}}  \\
\hline
RGHS \cite{RGHS}               & 13.68       & 12.91    & 14.20     & 13.84  & 11.64     & 13.66  & 11.08   & {\bf 13.00}      \\
\hline
ULAP \cite{ULAP}            & 20.98      & 17.06     & 20.36     & 15.56  & 24.47     & 20.09  & 20.53      & {\bf 19.86}        \\
\hline
Li et al. \cite{tip2016}             & 17.51       & 24.70    & 16.71     & 16.94  &  \textcolor{blue}{11.47}     & 13.48  & 13.93     & {\bf 16.39}     \\
\hline   
Ucolor \cite{Ucolor}             & \textcolor{blue}{10.73}      & 12.90    &\textcolor{blue}{11.30} & \textcolor{blue}{12.25}     & 14.34  & \textcolor{blue}{10.26}    & 10.79  & {\bf 11.79}          \\
\hline 
Waternet \cite{Waternet}            & 12.18    & 15.25   &11.66 & 15.74     & 19.89  & 12.14    & 11.31  & {\bf 14.02}           \\
\hline 
FUnIE-GAN \cite{EUVP}            & \textcolor{red}{10.29}     & 14.48    &12.67 & 13.81      & 15.91 & \textcolor{red}{10.11}     & 9.51  & {\bf 12.40}           \\
\hline 
UWCNN \cite{UWCNN}             & 12.97      & 13.29    &14.12 & 14.33     & 12.99 & 13.29     & 10.66 & {\bf 13.09}           \\
\hline 
USLN             & 10.88      & \textcolor{red}{10.69}    & \textcolor{red}{10.59} & 12.89      & \textcolor{red}{9.90}  & 11.00     & \textcolor{blue}{9.03}  & {\bf \textcolor{red}{10.71}}           \\    
\hline                         
\end{tabular}
\end{spacing} 
\label{tab3}
\vspace{-1em}
\end{table*}

\subsection{Visual Comparisons}
In this section, we conduct several visual comparisons on testing datasets. 

We first present raw images and enhanced images in paired datasets in Fig. \ref{UIEB}. Compared with other SOTA models, our result is closest to the ground-truth images and achieves the highest PSNR and SSIM scores. Note that, as table.1 shown, traditional methods get much lower PSNR/SSIM scores than deep-learning-model, so due to length limitation, we only present images enhanced by deep-learning-based models in Fig.7. 

The underwater images suffer from different color deviations are shown in Fig. \ref{UIEB}\subref{fig_first_case}. The first two images (left) are only distorted slightly, while the other images are more distorted. For slightly distorted images, all compared methods fail to restore the complete scene structure, these methods all get even lower PSNR or SSIM than raw images. In comparison, our model improves the quality of these images effectively without over-enhancement. For the rest images which are deviated by greenish and blueish color. The compared models either fail to remove the greenish and blueish color completely or introduce undesirable color artifacts. For example, these images enhanced by FUnIE-GAN and Ucolor are greenish and reddish, respectively. In comparison, our model recovers the color of image and achieves the best scores.

\begin{figure*}[!t]
\centering
\subfloat[Input]{\includegraphics[width=1.3in]{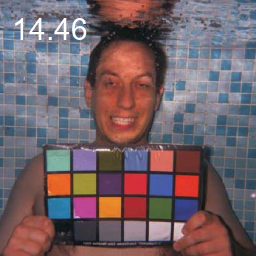}%
\label{fig_first_case}}
\hfil
\subfloat[UCM \cite{UCM}]{\includegraphics[width=1.3in]{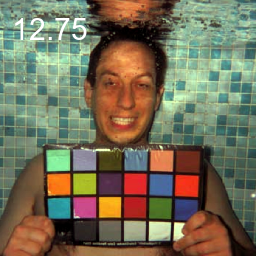}%
\label{fig_second_case}}
\hfil
\subfloat[RGHS \cite{RGHS}]{\includegraphics[width=1.3in]{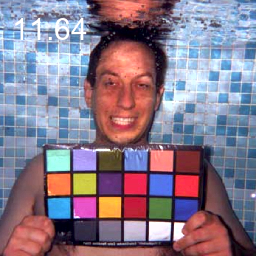}%
\label{fig_third_case}}
\hfil
\subfloat[ULAP \cite{ULAP}]{\includegraphics[width=1.3in]{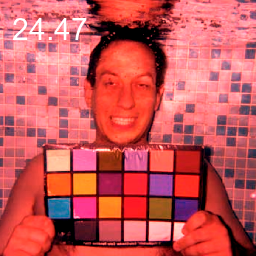}%
\label{fig_forth_case}}
\hfil
\subfloat[Li et al. \cite{tip2016}]{\includegraphics[width=1.3in]{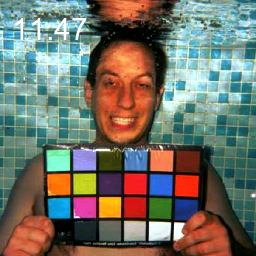}%
\label{fig_fifth_case}}
\hfil
\subfloat[Waternet \cite{Waternet}]{\includegraphics[width=1.3in]{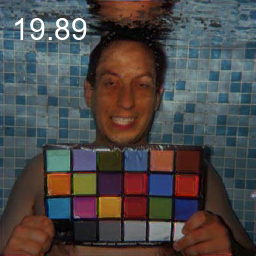}%
\label{fig_sixth_case}}
\hfil
\subfloat[FUnIE-GAN \cite{EUVP}]{\includegraphics[width=1.3in]{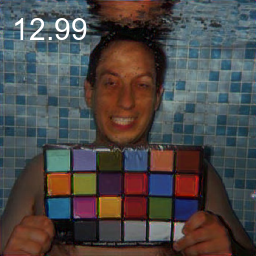}%
\label{fig_seventh_case}}
\hfil
\subfloat[UWCNN \cite{UWCNN}]{\includegraphics[width=1.3in]{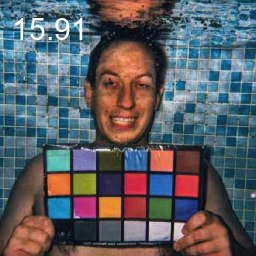}%
\label{fig_eighth_case}}
\hfil
\subfloat[Ucolor \cite{Ucolor}]{\includegraphics[width=1.3in]{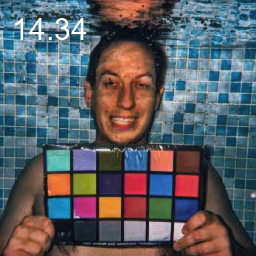}%
\label{fig_ninth_case}}
\hfil
\subfloat[Ours]{\includegraphics[width=1.3in]{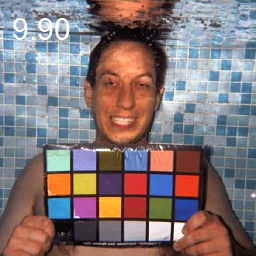}%
\label{fig_tenth_case}}
\caption{Visual comparisons on an underwater image taken by a Panasonic TS1 camera sampled from Color-Check7. Our model removes color cast and improves contrast effectively. Compared with the existing methods, the image produced by our USLN has the best score in terms of the CIEDE2000.}
\label{colorcheck}
\vspace{-1em}
\end{figure*}

\begin{figure*}[!t]
\centering
\subfloat[Input]{\includegraphics[width=1.1in]{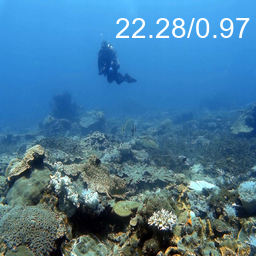}\ \includegraphics[width=1.1in]{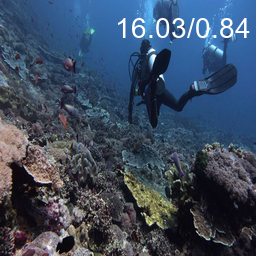}\ \includegraphics[width=1.1in]{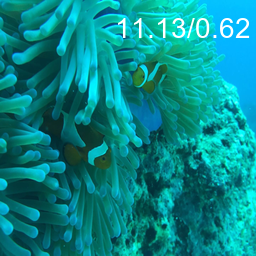}\ \includegraphics[width=1.1in]{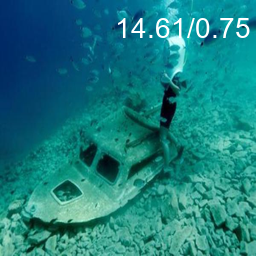}\ \includegraphics[width=1.1in]{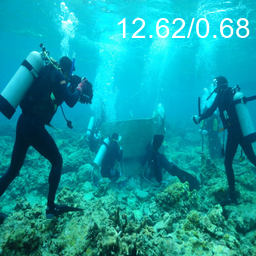}\ \includegraphics[width=1.1in]{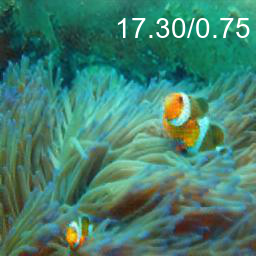}%
\label{fig_first_case}}
\hfil
\subfloat[Waternet \cite{Waternet}]{\includegraphics[width=1.1in]{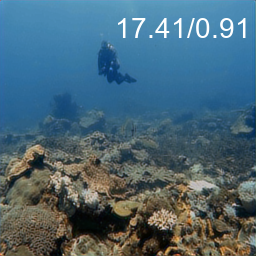}\ \includegraphics[width=1.1in]{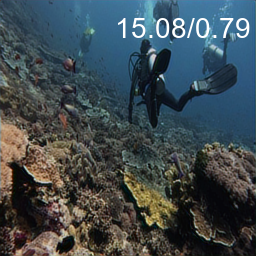}\ \includegraphics[width=1.1in]{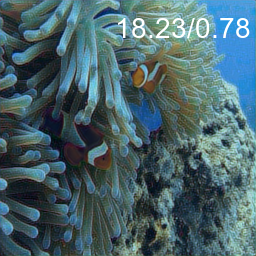}\ \includegraphics[width=1.1in]{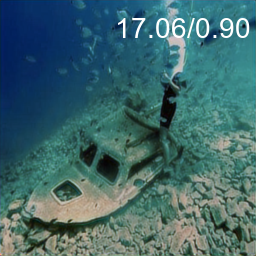}\ \includegraphics[width=1.1in]{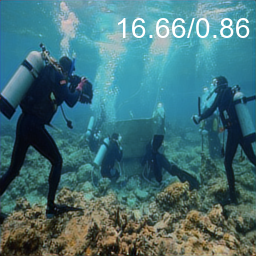}\ \includegraphics[width=1.1in]{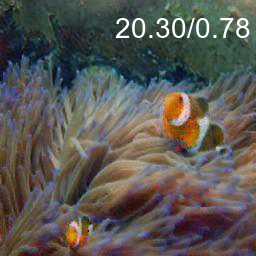}%
\label{fig_first_case}}
\hfil
\subfloat[UWCNN \cite{UWCNN}]{\includegraphics[width=1.1in]{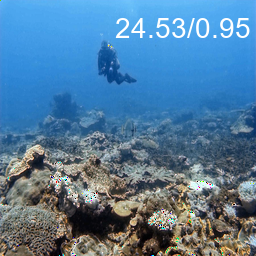}\ \includegraphics[width=1.1in]{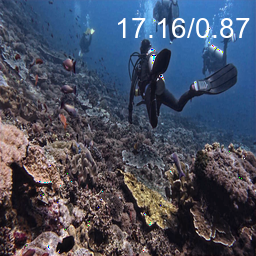}\ \includegraphics[width=1.1in]{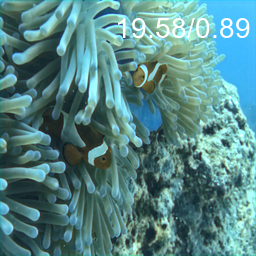}\ \includegraphics[width=1.1in]{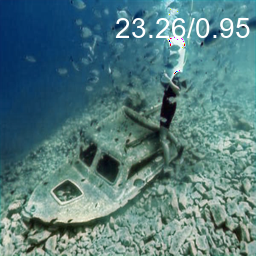}\ \includegraphics[width=1.1in]{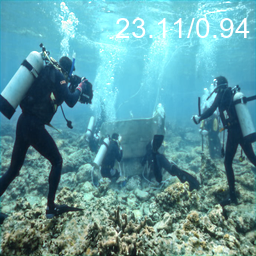}\ \includegraphics[width=1.1in]{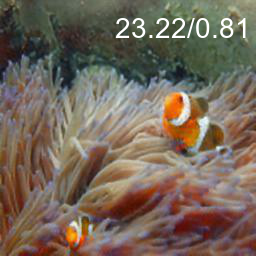}%
\label{fig_first_case}}
\hfil
\subfloat[Ucolor \cite{Ucolor}]{\includegraphics[width=1.1in]{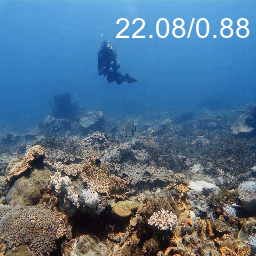}\ \includegraphics[width=1.1in]{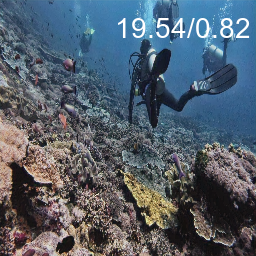}\ \includegraphics[width=1.1in]{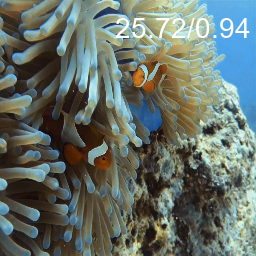}\ \includegraphics[width=1.1in]{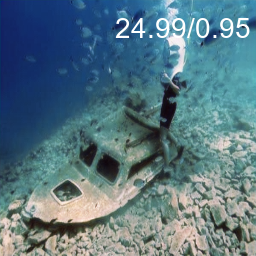}\ \includegraphics[width=1.1in]{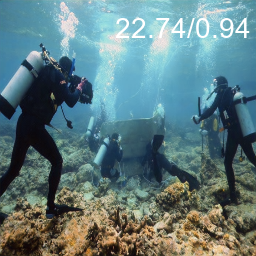}\ \includegraphics[width=1.1in]{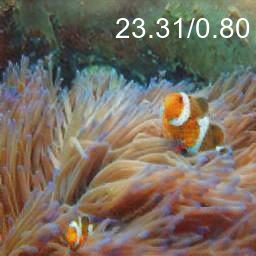}%
\label{fig_first_case}}
\hfil
\subfloat[FUnIE-GAN \cite{EUVP}]{\includegraphics[width=1.1in]{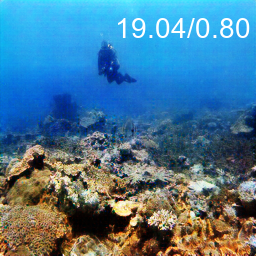}\ \includegraphics[width=1.1in]{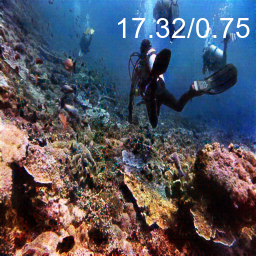}\ \includegraphics[width=1.1in]{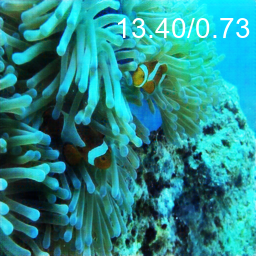}\ \includegraphics[width=1.1in]{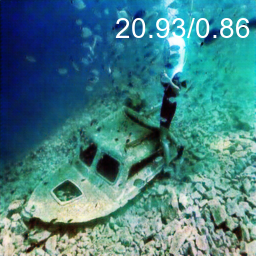}\ \includegraphics[width=1.1in]{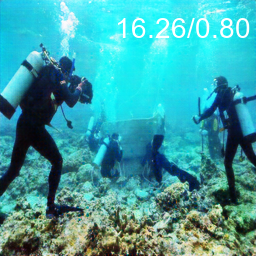}\ \includegraphics[width=1.1in]{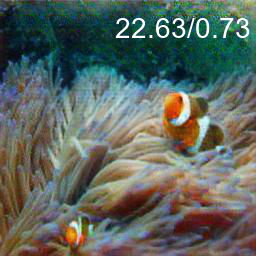}%
\label{fig_first_case}}
\hfil
\subfloat[USLN]{\includegraphics[width=1.1in]{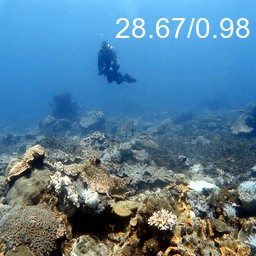}\ \includegraphics[width=1.1in]{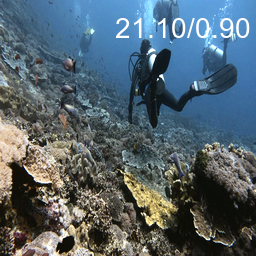}\ \includegraphics[width=1.1in]{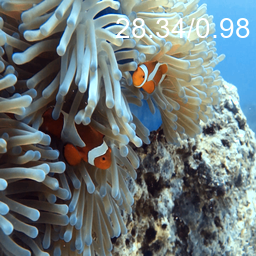}\ \includegraphics[width=1.1in]{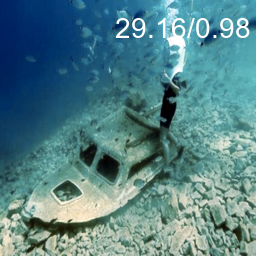}\ \includegraphics[width=1.1in]{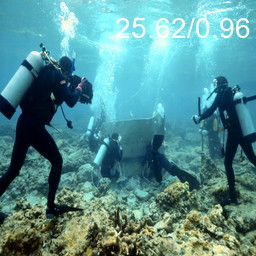}\ \includegraphics[width=1.1in]{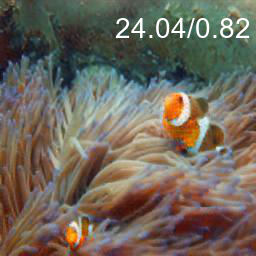}%
\label{fig_first_case}}
\caption{Visual comparisons on challenging underwater images sampled from Test-90 and Test-555. The number on the top-right corner of each image refers to its PSNR/SSIM (the larger, the better).}
\label{UIEB}
\vspace{-1em}
\end{figure*}

To demonstrate the robustness of our model, we then present the enhanced Color Checker images of different methods. The raw image shown in Fig. \ref{colorcheck} is taken by Panasonic TS1, which suffers from severe red color cast. All compared methods cannot achieve satisfactory results, some of them even reduce the quality of image. According to CIEDE2000 metric, our method gets the best score (9.90) and restores the most accurate colors of the raw image.

\begin{figure*}[!t]
\centering
\subfloat[Input]{\includegraphics[width=1.2in]{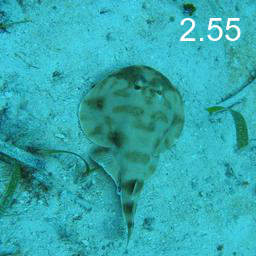}%
\label{fig_first_case}}
\hfil
\subfloat[UCM \cite{UCM}]{\includegraphics[width=1.2in]{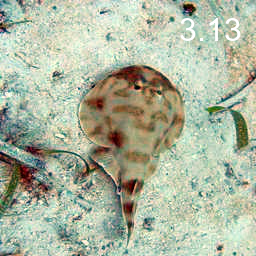}%
\label{fig_first_case}}
\hfil
\subfloat[RGHS \cite{RGHS}]{\includegraphics[width=1.2in]{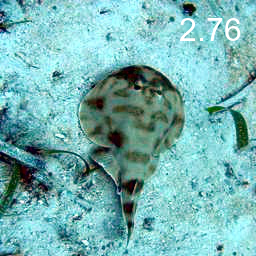}%
\label{fig_first_case}}
\hfil
\subfloat[ULAP \cite{ULAP}]{\includegraphics[width=1.2in]{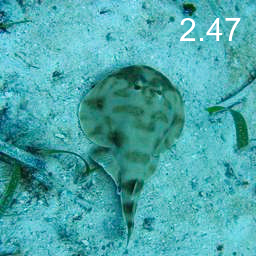}%
\label{fig_first_case}}
\hfil
\subfloat[Li et al. \cite{tip2016}]{\includegraphics[width=1.2in]{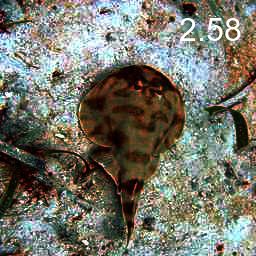}%
\label{fig_first_case}}
\hfil
\subfloat[Waternet \cite{Waternet}]{\includegraphics[width=1.2in]{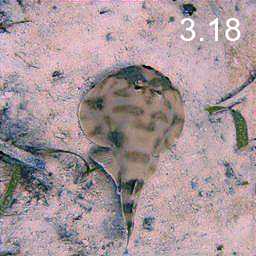}%
\label{fig_first_case}}
\hfil
\subfloat[UWCNN \cite{UWCNN}]{\includegraphics[width=1.2in]{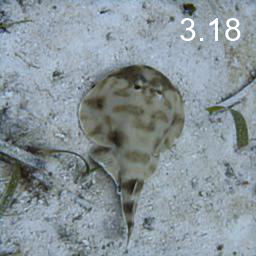}%
\label{fig_first_case}}
\hfil
\subfloat[FUnIE-GAN \cite{EUVP}]{\includegraphics[width=1.2in]{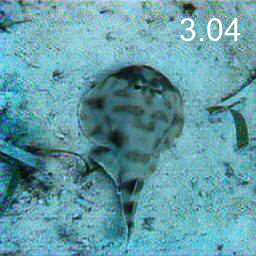}%
\label{fig_first_case}}
\hfil
\subfloat[Ucolor \cite{Ucolor}]{\includegraphics[width=1.2in]{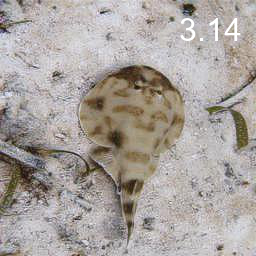}%
\label{fig_first_case}}
\hfil
\subfloat[USLN]{\includegraphics[width=1.2in]{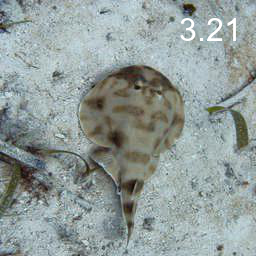}%
\label{fig_first_case}}
\caption{Visual comparisons on a real underwater image sampled from Test-1270. The numbers on the top-right corner of each image refer to its UIQM.}
\label{uiqm1}
\vspace{-1em}
\end{figure*}

\begin{figure*}[!t]
\centering
\subfloat[Input]{\includegraphics[width=1.2in]{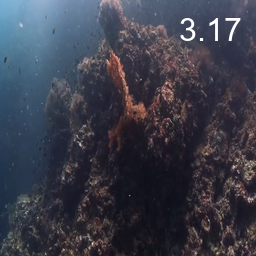}%
\label{fig_first_case}}
\hfil
\subfloat[UCM \cite{UCM}]{\includegraphics[width=1.2in]{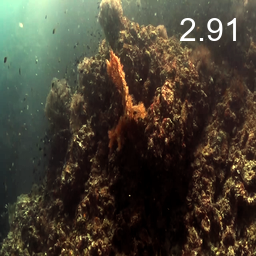}%
\label{fig_first_case}}
\hfil
\subfloat[RGHS \cite{RGHS}]{\includegraphics[width=1.2in]{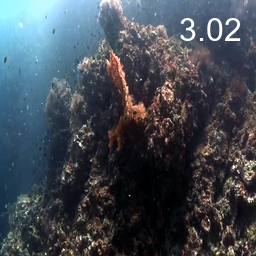}%
\label{fig_first_case}}
\hfil
\subfloat[ULAP \cite{ULAP}]{\includegraphics[width=1.2in]{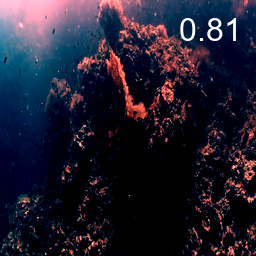}%
\label{fig_first_case}}
\hfil
\subfloat[Li et al. \cite{tip2016}]{\includegraphics[width=1.2in]{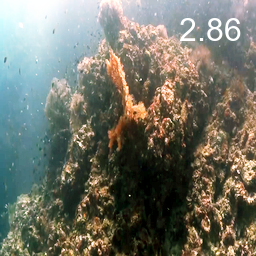}%
\label{fig_first_case}}
\hfil
\subfloat[Waternet \cite{Waternet}]{\includegraphics[width=1.2in]{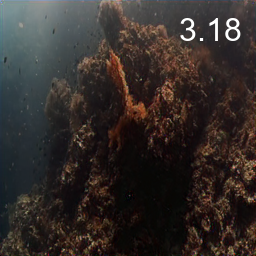}%
\label{fig_first_case}}
\hfil
\subfloat[UWCNN \cite{UWCNN}]{\includegraphics[width=1.2in]{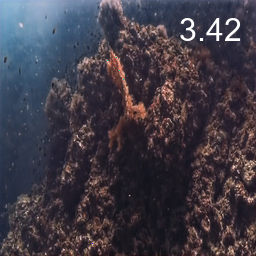}%
\label{fig_first_case}}
\hfil
\subfloat[FUnIE-GAN \cite{EUVP}]{\includegraphics[width=1.2in]{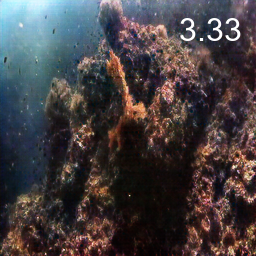}%
\label{fig_first_case}}
\hfil
\subfloat[Ucolor \cite{Ucolor}]{\includegraphics[width=1.2in]{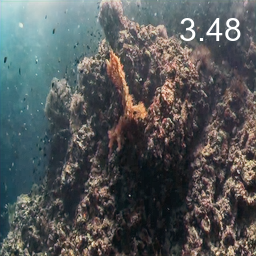}%
\label{fig_first_case}}
\hfil
\subfloat[USLN]{\includegraphics[width=1.2in]{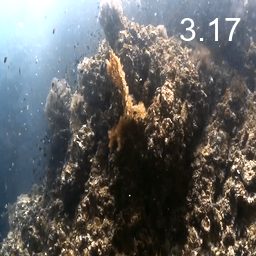}%
\label{fig_first_case}}
\caption{ Visual comparisons on a typical real underwater image with severe low-contrast sampled from Test-60. The numbers on the top-right corner of each image refer to its UIQM.}
\label{uiqm2}
\vspace{-1em}
\end{figure*}

\begin{figure*}[!t]
\centering
\subfloat[Input]{\includegraphics[width=1.2in]{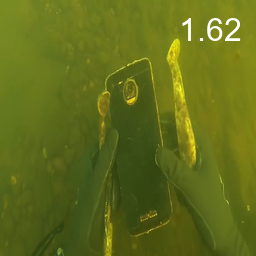}%
\label{input}}
\hfil
\subfloat[UCM \cite{UCM}]{\includegraphics[width=1.2in]{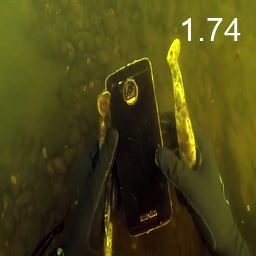}%
\label{fig_first_case}}
\hfil
\subfloat[RGHS \cite{RGHS}]{\includegraphics[width=1.2in]{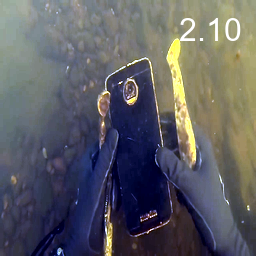}%
\label{fig_first_case}}
\hfil
\subfloat[ULAP \cite{ULAP}]{\includegraphics[width=1.2in]{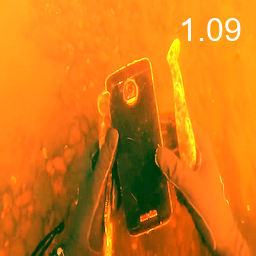}%
\label{fig_first_case}}
\hfil
\subfloat[Li et al. \cite{tip2016}]{\includegraphics[width=1.2in]{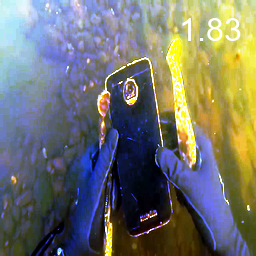}%
\label{fig_first_case}}
\hfil
\subfloat[Waternet \cite{Waternet}]{\includegraphics[width=1.2in]{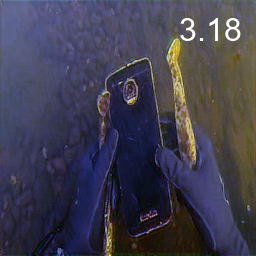}%
\label{waternet}}
\hfil
\subfloat[UWCNN \cite{UWCNN}]{\includegraphics[width=1.2in]{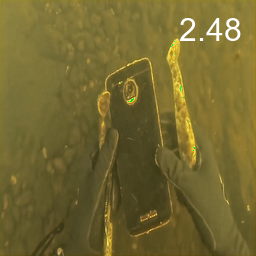}%
\label{fig_first_case}}
\hfil
\subfloat[FUnIE-GAN \cite{EUVP}]{\includegraphics[width=1.2in]{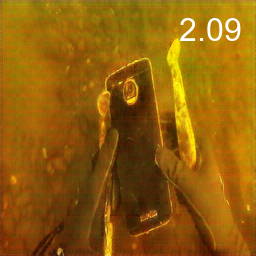}%
\label{fig_first_case}}
\hfil
\subfloat[Ucolor \cite{Ucolor}]{\includegraphics[width=1.2in]{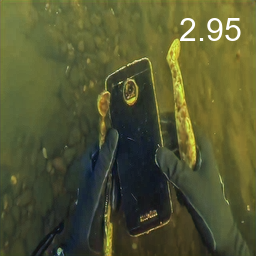}%
\label{fig_first_case}}
\hfil
\subfloat[USLN]{\includegraphics[width=1.2in]{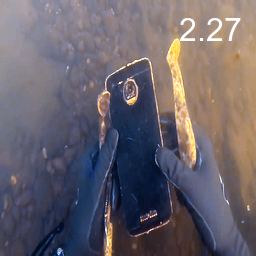}%
\label{fig_first_case}}
\caption{Visual comparisons on a typical real underwater image with obvious yellow color deviation sampled from Test-60. The numbers on the top-right corner of each image refer to its UIQM.}
\label{uiqm3}
\vspace{-1em}
\end{figure*}

We then show raw images and enhanced images in unpaired datasets, Test-60 and Test-1270 in Fig. \ref{uiqm1}, Fig. \ref{uiqm2}, and Fig. \ref{uiqm3}. Noted that we only use UIQM as the reference to evaluate these images, the accurate perceptual quality of these enhanced images need the readers to score.

For Fig. \ref{uiqm1}, the raw image sampled from Test-1270 suffers from serious blueish deviation. Our result gets the best UIQM. Compared with SOTA models, our model successes to recover the natural color and structural details. 

For Fig. \ref{uiqm2} and Fig. \ref{uiqm3} which are sampled from Test-60, these two images are much more challenging. As Fig. \ref{uiqm2} shown, the obvious light limitation of the raw image results in low contrast. The compared models tend to make the image darker, some models even introduce reddish color. In comparison, our model increases both brightness and contrast, making the details of the image clear. Fig. \ref{uiqm3}\subref{input} suffers from obvious yellow deviation, all compared methods cannot achieve satisfactory results. For example, although Fig. \ref{uiqm3}\subref{waternet} which is enhanced by Waternet achieves the best UIQM, it is obvious that the left part of the image is over-enhanced, introducing unwanted blue, while the right part is under-enhanced, failing to remove yellow completely. In comparison, our model removes the yellowish color and makes the color balanced.

\subsection{Ablation}
We perform extensive ablation studies to analyze the advantage of the core modules, including residual-enhancement module (REM), multi-color space stretch module (MCSM), and dual-statistic white balance module (DSBM). In addition, we analyze the combined loss in our model. More specifically,
\begin{itemize}
\item{w/o Lab+SI, w/o Lab, and w/o SI mean that the multi-color space stretch module without both Lab and SI color channels, Lab channels, and SI channels, respectively. w/ H means that we do not separate H channel out in multi-color space stretch module.}
\item{w/o avg and w/o max, and w/o max+avg stand for dual-statistic white balance module without average, maximum, and both average and maximum, respectively.}
\item{w/o RES refers to USLN without residual-enhancement module, while w/o Tanh refers to the Residual-Enhancement Module without Tanh activation function.}
\item{w/o VGG+SSIM, w/o SSIM, and w/o VGG mean that USLN is trained with MAE loss, MAE+SSIM loss, and MAE+VGG loss, respectively.}
\end{itemize}

\begin{figure}[!t]
\centering
\subfloat[Input]{\includegraphics[width=1.1in]{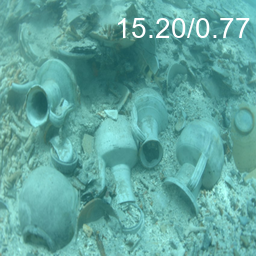}%
\label{fig_first_case}}
\hfil
\subfloat[w/o max+avg]{\includegraphics[width=1.1in]{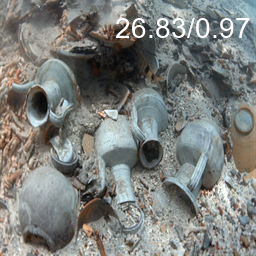}%
\label{fig_second_case}}
\hfil
\subfloat[w/o max]{\includegraphics[width=1.1in]{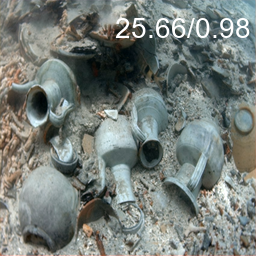}%
\label{fig_third_case}}
\hfil
\subfloat[w/o avg]{\includegraphics[width=1.1in]{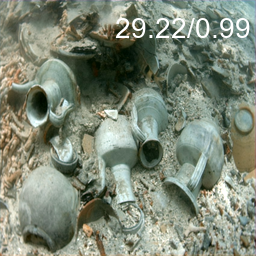}%
\label{fig_fourth_case}}
\hfil
\subfloat[USLN]{\includegraphics[width=1.1in]{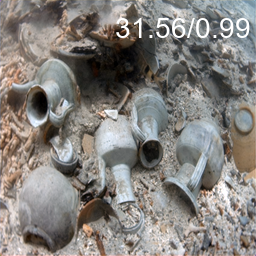}%
\label{fig_fourth_case}}
\hfil
\subfloat[label]{\includegraphics[width=1.1in]{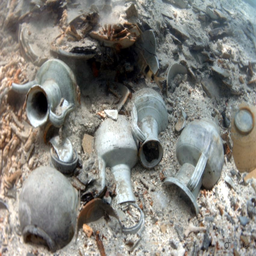}%
\label{fig_fourth_case}}
\caption{Ablation study of the contributions of dual-statistic white balance module. The ablated models cannot remove the color casts completely, which implies average and maximum of image is both useful for model to balance the color channrls. The number on the top-right corner of each image refers to its PSNR/SSIM (the larger, the better).}
\label{dual}
\vspace{-1em}
\end{figure}

\begin{figure}[!t]
\centering
\subfloat[Input]{\includegraphics[width=1.1in]{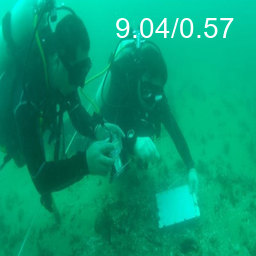}%
\label{fig_first_case}}
\hfil
\subfloat[w/o Lab+SI]{\includegraphics[width=1.1in]{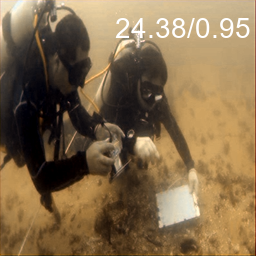}%
\label{fig_second_case}}
\hfil
\subfloat[w/o Lab]{\includegraphics[width=1.1in]{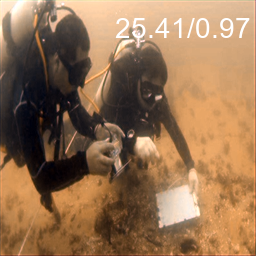}%
\label{fig_third_case}}
\hfil
\subfloat[w/o SI]{\includegraphics[width=1.1in]{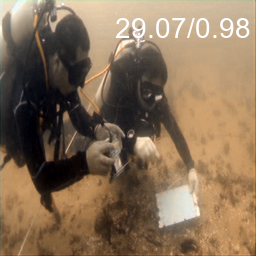}%
\label{fig_fourth_case}}
\hfil
\subfloat[USLN]{\includegraphics[width=1.1in]{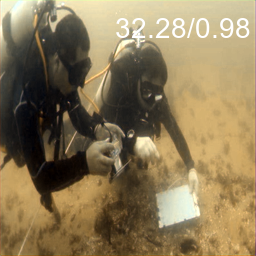}%
\label{fig_fourth_case}}
\hfil
\subfloat[label]{\includegraphics[width=1.1in]{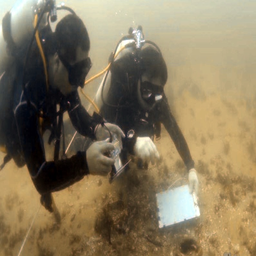}%
\label{fig_fourth_case}}
\caption{Ablation study of the contributions of multi-color space stretch module. The full model which enhances image in three colors can restore the inherent color. The number on the top-right corner of each image refers to its PSNR/SSIM (the larger, the better).}
\label{multi}
\vspace{-1em}
\end{figure}

\begin{figure}[!t]
\centering
\subfloat[Input]{\includegraphics[width=1.1in]{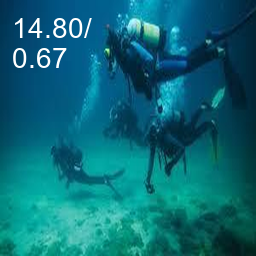}%
\label{fig_first_case}}
\hfil
\subfloat[w/o Res]{\includegraphics[width=1.1in]{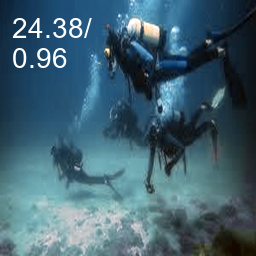}%
\label{fig_second_case}}
\hfil
\subfloat[USLN]{\includegraphics[width=1.1in]{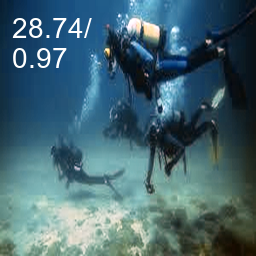}%
\label{fig_third_case}}
\caption{Ablation study of the contributions of residual-enhancement module. residual-Enhancement module can preserve the image fidelity, making the enhanced images look more natural. The number on the top-left corner of each image refers to its PSNR/SSIM (the larger, the better).}
\label{resa}
\vspace{-1em}
\end{figure}

We used MSE, PSNR, and SSIM to evaluate the results on Test-90, which are shown in Table \uppercase\expandafter{\romannumeral6}. Compared to the ablation models, full model achieves the best quantitative performance on all metrics. The visual comparisons of the contributions of dual-statistic white balance module, muti-color stretch model, and residual-enhancement model are presented in Fig. \ref{dual}, Fig. \ref{multi}, and Fig. \ref{resa}, respectively. The conclusions drawn from the ablation study for the each module are listed as follows:

1)The ablated models w/o avg and w/o max produce comparable performance as shown in Table \uppercase\expandafter{\romannumeral7} and Fig. \ref{dual}. Dual-statistic balance module help model to balance the images base on two representative values, average and maximum, while removing either of them will degrade performance as Table \uppercase\expandafter{\romannumeral7} shown. As Fig. \ref{dual} shown, the full model removes the color casts much more completely than ablated models.

2)As Table \uppercase\expandafter{\romannumeral7} shown, the lack of stretch in either HSI or Lab space will decrease performance. In addition, as Fig. \ref{multi} shown, the full model enhances images from three color spaces, which is effective to restore the natural color of image. Interestingly, we find blindly add all channels in HSI color space will reduce the performance, while adding only S and I channels can improve performance effectively.

3)Residual-enhancement module can preserve the image fidelity, making the enhanced images look more natural. As presented in Fig. \ref{resa}, w/o Res model over-enhances the image, making aquatic weeds be greyish, while full model restores normal green color of aquatic weeds. Moreover, the lack of Tanh activation function will degrade the performance as Table \uppercase\expandafter{\romannumeral7} presented.

4)Ablation study towards the combined loss is presented in Table \uppercase\expandafter{\romannumeral7}. SSIM loss in the combined loss function can improve the structural similarity of the output image as shown in Table \uppercase\expandafter{\romannumeral7}, while VGG perceptual loss can improve the visual quality of output. By adding perceptual loss and SSIM loss to the MAE loss, the visual quality of final result is improved.

\begin{table}
\centering
\caption{QUANTITATIVE RESULTS OF THE ABLATION STUDY IN TERMS OF AVERAGE MSE, PSNR AND SSIM VALUES}
\label{table}
\small
\begin{spacing}{1.19}
\setlength{\tabcolsep}{7pt}
\begin{tabular}{ccccc}
\hline
Modules                       & Baselines  & MSE $\downarrow$ & PSNR $\uparrow$ & SSIM $\uparrow$ \\
\hline
                              & full model & 479   & 23.78    & 0.92    \\
\hline
\multirow{3}{*}{MCSM}    & w/o Lab+SI        & 540   & 23.38    & 0.91    \\
                              & w/o Lab    & 505   & 23.51    & 0.91    \\
                              & w/o SI    & 494   & 23.57    & 0.92    \\
							  & w/ HSI    & 486   & 23.48    & 0.91    \\
\hline
\multirow{3}{*}{DSBM} & w/o max         & 556   & 22.98    & 0.91    \\
                              & w/o avg         & 679   & 22.33    & 0.90    \\
							  & w/o max+avg     & 763   & 21.20    & 0.90    \\
\hline
\multirow{2}{*}{REM}       & w/o RES     & 611   & 22.97    & 0.91    \\
							  & w/o Tanh     & 506   & 23.51    & 0.91    \\
\hline
\multirow{3}{*}{LOSS}         & w/o VGG+SSIM        & 491   & 23.38    & 0.91    \\
                              & w/o VGG   & 470   & 23.73    & 0.92    \\
                              & w/o SSIM    & 493   & 23.77    & 0.91    \\
\hline
\end{tabular}
\end{spacing}
\label{tab4}
\vspace{-1em}
\end{table}

\subsection{Failure Case}

\begin{figure}[!t]
\centering
\subfloat[Input]{\includegraphics[width=1.1in]{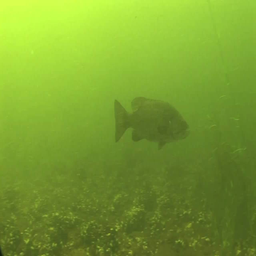}%
\label{fig_first_case}}
\hfil
\subfloat[Ucolor \cite{Ucolor}]{\includegraphics[width=1.1in]{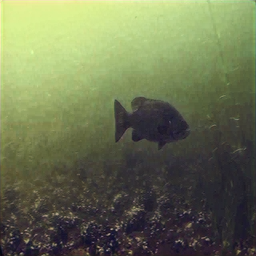}%
\label{fig_second_case}}
\hfil
\subfloat[USLN]{\includegraphics[width=1.1in]{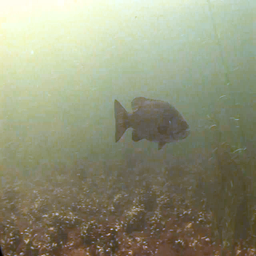}%
\label{fig_third_case}}
\caption{Failure case. The input underwater image suffers from severe yellow color cast. USLN removes the yellow color but introduces unwanted red on the sands.}
\label{fail}
\vspace{-1em}
\end{figure}
Enhancing the underwater image with severe color cast is very challenging. Both our model and other SOTA models fail to work well. As Fig. \ref{fail} shown, although our model and Ucolor remove the yellow cast, it is far from achieving the desired visual quality. Ucolor makes image much darker, while USLN introduces unwanted red. The main reason is the lack of similar images in train sets, limiting the ability of deep-learning-based methods to deal with such images.

\section{Conclusion}
In this paper, we introduce a statistically guided lightweight networks for underwater image enhancement via dual-statistic white balance and multi-color space stretch. Our model introduces several statistic-based methods into networks, including Gray World, White Patch and histogram stretching. Our model aims to enhance images by adjusting the pixel values directly, instead of processing high-dimensional image features. As a result, our model takes only 894 trainable parameters, but provides state-of-the-art performance in diverse representative datasets(UIEB, EUVP, UFO). We also conduct extensive ablation experiments to verity the effectiveness of the main components.

\section*{Acknowledgments}
The authors would like to thank Chongyi Li, Chunle Guo, Wenqi Ren, Runmin Cong, Junhui Hou, Sam Kwong, and Dacheng Tao for providing the UIEB dataset. They would also like to thank Md Jahidul Islam, Youya Xia, Peigen Luo and Junaed Sattar for providing EUVP and UFO datasets. Finally, they would also like to thank Codruta O. Ancuti, Cosmin Ancuti, Christophe De Vleeschouwer, and Philippe Bekaert for providing the underwater Color Checker images.

\small
\bibliographystyle{IEEEtran}
\bibliography{reference}

\begin{thebibliography}{10}
\providecommand{\url}[1]{#1}
\csname url@samestyle\endcsname
\providecommand{\newblock}{\relax}
\providecommand{\bibinfo}[2]{#2}
\providecommand{\BIBentrySTDinterwordspacing}{\spaceskip=0pt\relax}
\providecommand{\BIBentryALTinterwordstretchfactor}{4}
\providecommand{\BIBentryALTinterwordspacing}{\spaceskip=\fontdimen2\font plus
\BIBentryALTinterwordstretchfactor\fontdimen3\font minus
  \fontdimen4\font\relax}
\providecommand{\BIBforeignlanguage}[2]{{%
\expandafter\ifx\csname l@#1\endcsname\relax
\typeout{** WARNING: IEEEtran.bst: No hyphenation pattern has been}%
\typeout{** loaded for the language `#1'. Using the pattern for}%
\typeout{** the default language instead.}%
\else
\language=\csname l@#1\endcsname
\fi
#2}}
\providecommand{\BIBdecl}{\relax}
\BIBdecl

\bibitem{zhang2021enhancing}
W.~Zhang, L.~Dong, T.~Zhang, and W.~Xu, ``Enhancing underwater image via color
  correction and bi-interval contrast enhancement,'' \emph{Signal Processing:
  Image Communication}, vol.~90, p. 116030, 2021.

\bibitem{liu2020semantic}
F.~Liu and M.~Fang, ``Semantic segmentation of underwater images based on
  improved deeplab,'' \emph{Journal of Marine Science and Engineering}, vol.~8,
  no.~3, p. 188, 2020.

\bibitem{UCM}
K.~Iqbal, M.~Odetayo, A.~James, R.~A. Salam, and A.~Z.~H. Talib, ``Enhancing
  the low quality images using unsupervised colour correction method,'' in
  \emph{2010 IEEE International Conference on Systems, Man and
  Cybernetics}.\hskip 1em plus 0.5em minus 0.4em\relax IEEE, 2010, pp.
  1703--1709.

\bibitem{RGHS}
D.~Huang, Y.~Wang, W.~Song, J.~Sequeira, and S.~Mavromatis, ``Shallow-water
  image enhancement using relative global histogram stretching based on
  adaptive parameter acquisition,'' in \emph{International conference on
  multimedia modeling}.\hskip 1em plus 0.5em minus 0.4em\relax Springer, 2018,
  pp. 453--465.

\bibitem{ULAP}
W.~Song, Y.~Wang, D.~Huang, and D.~Tjondronegoro, ``A rapid scene depth
  estimation model based on underwater light attenuation prior for underwater
  image restoration,'' in \emph{Pacific Rim Conference on Multimedia}.\hskip
  1em plus 0.5em minus 0.4em\relax Springer, 2018, pp. 678--688.

\bibitem{Waternet}
C.~Li, C.~Guo, W.~Ren, R.~Cong, J.~Hou, S.~Kwong, and D.~Tao, ``An underwater
  image enhancement benchmark dataset and beyond,'' \emph{IEEE Transactions on
  Image Processing}, vol.~29, pp. 4376--4389, 2019.

\bibitem{UWCNN}
C.~Li, S.~Anwar, and F.~Porikli, ``Underwater scene prior inspired deep
  underwater image and video enhancement,'' \emph{Pattern Recognition},
  vol.~98, p. 107038, 2020.

\bibitem{Ucolor}
C.~Li, S.~Anwar, J.~Hou, R.~Cong, C.~Guo, and W.~Ren, ``Underwater image
  enhancement via medium transmission-guided multi-color space embedding,''
  \emph{IEEE Transactions on Image Processing}, vol.~30, pp. 4985--5000, 2021.

\bibitem{EUVP}
M.~J. Islam, Y.~Xia, and J.~Sattar, ``Fast underwater image enhancement for
  improved visual perception,'' \emph{IEEE Robotics and Automation Letters},
  vol.~5, no.~2, pp. 3227--3234, 2020.

\bibitem{ARUIR}
A.~Galdran, D.~Pardo, A.~Pic{\'o}n, and A.~Alvarez-Gila, ``Automatic
  red-channel underwater image restoration,'' \emph{Journal of Visual
  Communication and Image Representation}, vol.~26, pp. 132--145, 2015.

\bibitem{UDCP}
P.~Drews, E.~Nascimento, F.~Moraes, S.~Botelho, and M.~Campos, ``Transmission
  estimation in underwater single images,'' in \emph{Proceedings of the IEEE
  international conference on computer vision workshops}, 2013, pp. 825--830.

\bibitem{tip2016}
C.-Y. Li, J.-C. Guo, R.-M. Cong, Y.-W. Pang, and B.~Wang, ``Underwater image
  enhancement by dehazing with minimum information loss and histogram
  distribution prior,'' \emph{IEEE Transactions on Image Processing}, vol.~25,
  no.~12, pp. 5664--5677, 2016.

\bibitem{ICM}
K.~Iqbal, R.~A. Salam, A.~Osman, and A.~Z. Talib, ``Underwater image
  enhancement using an integrated colour model.'' \emph{IAENG International
  Journal of computer science}, vol.~34, no.~2, 2007.

\bibitem{ghani2015enhancement}
A.~S.~A. Ghani and N.~A.~M. Isa, ``Enhancement of low quality underwater image
  through integrated global and local contrast correction,'' \emph{Applied Soft
  Computing}, vol.~37, pp. 332--344, 2015.

\bibitem{ghani2017automatic}
------, ``Automatic system for improving underwater image contrast and color
  through recursive adaptive histogram modification,'' \emph{Computers and
  electronics in agriculture}, vol. 141, pp. 181--195, 2017.

\bibitem{liu2020real}
R.~Liu, X.~Fan, M.~Zhu, M.~Hou, and Z.~Luo, ``Real-world underwater
  enhancement: Challenges, benchmarks, and solutions under natural light,''
  \emph{IEEE Transactions on Circuits and Systems for Video Technology},
  vol.~30, no.~12, pp. 4861--4875, 2020.

\bibitem{li2020hybrid}
X.~Li, G.~Hou, L.~Tan, and W.~Liu, ``A hybrid framework for underwater image
  enhancement,'' \emph{IEEE Access}, vol.~8, pp. 197\,448--197\,462, 2020.

\bibitem{finlayson2004shades}
G.~D. Finlayson and E.~Trezzi, ``Shades of gray and colour constancy,'' in
  \emph{Color and Imaging Conference}, vol. 2004, no.~1.\hskip 1em plus 0.5em
  minus 0.4em\relax Society for Imaging Science and Technology, 2004, pp.
  37--41.

\bibitem{CLAHE}
K.~Zuiderveld, ``Contrast limited adaptive histogram equalization,''
  \emph{Graphics gems}, pp. 474--485, 1994.

\bibitem{pixelpixel}
X.~Sun, L.~Liu, Q.~Li, J.~Dong, E.~Lima, and R.~Yin, ``Deep pixel-to-pixel
  network for underwater image enhancement and restoration,'' \emph{IET Image
  Processing}, vol.~13, no.~3, pp. 469--474, 2019.

\bibitem{DCP}
K.~He, J.~Sun, and X.~Tang, ``Single image haze removal using dark channel
  prior,'' \emph{IEEE transactions on pattern analysis and machine
  intelligence}, vol.~33, no.~12, pp. 2341--2353, 2010.

\bibitem{CNN}
Y.~LeCun, L.~Bottou, Y.~Bengio, and P.~Haffner, ``Gradient-based learning
  applied to document recognition,'' \emph{Proceedings of the IEEE}, vol.~86,
  no.~11, pp. 2278--2324, 1998.

\bibitem{GAN}
I.~Goodfellow, J.~Pouget-Abadie, M.~Mirza, B.~Xu, D.~Warde-Farley, S.~Ozair,
  A.~Courville, and Y.~Bengio, ``Generative adversarial nets,'' \emph{Advances
  in neural information processing systems}, vol.~27, 2014.

\bibitem{unet}
O.~Ronneberger, P.~Fischer, and T.~Brox, ``U-net: Convolutional networks for
  biomedical image segmentation,'' in \emph{International Conference on Medical
  image computing and computer-assisted intervention}.\hskip 1em plus 0.5em
  minus 0.4em\relax Springer, 2015, pp. 234--241.

\bibitem{shallow}
A.~Naik, A.~Swarnakar, and K.~Mittal, ``Shallow-uwnet: Compressed model for
  underwater image enhancement (student abstract),'' in \emph{Proceedings of
  the AAAI Conference on Artificial Intelligence}, vol.~35, no.~18, 2021, pp.
  15\,853--15\,854.

\bibitem{naik2003hue}
S.~K. Naik and C.~Murthy, ``Hue-preserving color image enhancement without
  gamut problem,'' \emph{IEEE Transactions on image processing}, vol.~12,
  no.~12, pp. 1591--1598, 2003.

\bibitem{vgg}
K.~Simonyan and A.~Zisserman, ``Very deep convolutional networks for
  large-scale image recognition,'' \emph{arXiv preprint arXiv:1409.1556}, 2014.

\bibitem{abdul2014underwater}
A.~S. Abdul~Ghani and N.~A. Mat~Isa, ``Underwater image quality enhancement
  through composition of dual-intensity images and rayleigh-stretching,''
  \emph{SpringerPlus}, vol.~3, no.~1, pp. 1--14, 2014.

\bibitem{res}
K.~He, X.~Zhang, S.~Ren, and J.~Sun, ``Deep residual learning for image
  recognition,'' in \emph{Proceedings of the IEEE conference on computer vision
  and pattern recognition}, 2016, pp. 770--778.

\bibitem{zhao2016loss}
H.~Zhao, O.~Gallo, I.~Frosio, and J.~Kautz, ``Loss functions for image
  restoration with neural networks,'' \emph{IEEE Transactions on computational
  imaging}, vol.~3, no.~1, pp. 47--57, 2016.

\bibitem{johnson2016perceptual}
J.~Johnson, A.~Alahi, and L.~Fei-Fei, ``Perceptual losses for real-time style
  transfer and super-resolution,'' in \emph{European conference on computer
  vision}.\hskip 1em plus 0.5em minus 0.4em\relax Springer, 2016, pp. 694--711.

\bibitem{SE}
J.~Hu, L.~Shen, and G.~Sun, ``Squeeze-and-excitation networks,'' in
  \emph{Proceedings of the IEEE conference on computer vision and pattern
  recognition}, 2018, pp. 7132--7141.

\bibitem{UFO}
M.~J. Islam, P.~Luo, and J.~Sattar, ``Simultaneous enhancement and
  super-resolution of underwater imagery for improved visual perception,''
  \emph{arXiv preprint arXiv:2002.01155}, 2020.

\bibitem{colorcheck}
C.~O. Ancuti, C.~Ancuti, C.~De~Vleeschouwer, and P.~Bekaert, ``Color balance
  and fusion for underwater image enhancement,'' \emph{IEEE Transactions on
  image processing}, vol.~27, no.~1, pp. 379--393, 2017.

\bibitem{psnrssim}
Z.~Wang, A.~C. Bovik, H.~R. Sheikh, and E.~P. Simoncelli, ``Image quality
  assessment: from error visibility to structural similarity,'' \emph{IEEE
  transactions on image processing}, vol.~13, no.~4, pp. 600--612, 2004.

\bibitem{UIQM}
K.~Panetta, C.~Gao, and S.~Agaian, ``Human-visual-system-inspired underwater
  image quality measures,'' \emph{IEEE Journal of Oceanic Engineering},
  vol.~41, no.~3, pp. 541--551, 2015.

\bibitem{berman2020underwater}
D.~Berman, D.~Levy, S.~Avidan, and T.~Treibitz, ``Underwater single image color
  restoration using haze-lines and a new quantitative dataset,'' \emph{IEEE
  transactions on pattern analysis and machine intelligence}, vol.~43, no.~8,
  pp. 2822--2837, 2020.

\bibitem{CIEDE}
G.~Sharma, W.~Wu, and E.~N. Dalal, ``The ciede2000 color-difference formula:
  Implementation notes, supplementary test data, and mathematical
  observations,'' \emph{Color Research \& Application: Endorsed by
  Inter-Society Color Council, The Colour Group (Great Britain), Canadian
  Society for Color, Color Science Association of Japan, Dutch Society for the
  Study of Color, The Swedish Colour Centre Foundation, Colour Society of
  Australia, Centre Fran{\c{c}}ais de la Couleur}, vol.~30, no.~1, pp. 21--30,
  2005.

\end{thebibliography}

\vfill

\end{document}